# Multi-Model Ensemble and Reservoir Computing for River Discharge Prediction in Ungauged Basins

Mizuki Funato[1] & Yohei Sawada[1]

[1]Department of Civil Engineering, The University of Tokyo, Tokyo, Japan

*Correspondence to*: Mizuki Funato (mizuki-funato@g.ecc.u-tokyo.ac.jp)

**Key Points**

- Combining multi-model ensembles with Reservoir Computing yields an accurate, interpretable, and efficient model for data-scarce regions
- Prediction in ungauged basins is enabled by linking trained model weights to physical basin attributes, creating a generalizable framework
- Predictive limitations of individual uncalibrated conceptual hydrological models can be mitigated with machine learning and ensemble methods

**Abstract.** Despite the necessity for accurate flood prediction, many regions lack sufficient river discharge observations. Although numerous models for daily river discharge prediction exist, achieving high accuracy, interpretability, and efficiency under data-scarce conditions remains a major challenge. We address this with a novel method, HYdrological Prediction with multi-model Ensemble and Reservoir computing (HYPER). Our approach applies Bayesian model averaging (BMA) to 47 "uncalibrated" catchment-based conceptual hydrological models. A reservoir computing (RC) model, a type of machine learning model, is then trained via linear regression to correct BMA output errors, a non-iterative process ensuring computational efficiency. For ungauged basins, we infer the required BMA and RC weights by mapping them to catchment attributes from gauged basins, creating a generalizable framework. Evaluated on 87 Japanese basins, in a data-rich scenario, HYPER (median Nash Sutcliffe Efficiency, NSE, of 0.59) performed comparably to a benchmark LSTM (NSE 0.64) but required only 3 % of its computational time. In a data-scarce scenario (where only ~20 % of basins are gauged), HYPER maintained robust performance (NSE 0.51) by leveraging the physical structure of the ensemble. In contrast, the LSTM's performance degraded substantially (NSE -0.61) due to data insufficiency. These results demonstrate that calibrating individual conceptual hydrological models is unnecessary when using a sufficiently large ensemble that is assembled and combined with machine-learning-based bias correction. HYPER provides a robust, efficient, and generalizable solution for discharge prediction, particularly in ungauged basins. By eliminating basin-specific calibration, HYPER offers a scalable, interpretable framework for accurate hydrological prediction in diverse data-scarce regions.

**Plain Language Summary**

By combining a multi-model ensemble of conceptual hydrological models with reservoir computing, we developed a highly accurate, interpretable, and efficient model for daily discharge prediction that works well with limited discharge data. Our method shows strong performance in both gauged and ungauged basins, remains robust in data-scarce scenarios, requires no iterative calculations, and provides insights into underlying hydrological processes. This offers a scalable alternative to conventional and deep learning models.

## 1 Introduction

Accurate river discharge observation in daily time steps is essential for effective flood prediction and prevention, water resource management, and climate change adaptation. While the Global Runoff Data Centre has recorded 10,831 river gauge stations worldwide, approximately 37% of these became inactive by 2020 (Global Runoff Data Centre, 2025). Moreover, active stations with discharge observations between 2021 and 2024 are predominantly located in North America, Europe, and Australia. Extensive regions across Asia, Africa, and South America face a significant lack of river discharge data, despite often being the most vulnerable to challenges such as flooding and climate change-induced risks (Li et al., 2019). This data scarcity highlights the need for alternative methods to estimate river discharge in ungauged river basins.

When river discharge data are unavailable, predictive models are essential for estimation. These models can be broadly categorized into hydrological models and machine learning models. Hydrological models rely on governing equations that capture the physical and empirical relationships between meteorological conditions (e.g., precipitation, temperature) and runoff. They are further divided into physically based models and conceptual models. Physically based models (e.g., Dai et al., 2003; Takata et al., 2003) describe hydrological processes using governing equations derived from solving the fundamental laws of physics such as conservation of mass, momentum, and energy between land and atmosphere to simulate physical processes in soil and vegetation. While they can theoretically be applied to ungauged basins by using measurable catchment characteristics, they demand significant computational resources and detailed input data. In contrast, conceptual models (e.g., Bergström, 1995; Ye et al., 2012; Moore, 1985) use simplified concepts such as buckets to approximate the hydrological processes by focusing on key dynamics. Although they also rely on equations, these rely on empirical relationships rather than fundamental physics, and thus, generally require extensive parameter calibration. The requirement for extensive parameter calibration complicates their applications in ungauged river basins. Extensive research has explored parameter estimation methods for conceptual models toward their applications in ungauged river basins, including spatial proximity, physical proximity, and regression-based approaches (Beck et al., 2016; He et al., 2011; Kokkonen et al., 2003; Yang et al., 2018; Zhang et al., 2008). The spatial proximity method, which selects the nearest gauged basin as the donor, is valued for its simplicity and accessibility. However, it often overlooks local variations, which can decrease performance, particularly in regions where gauged basins are scarce (Parajka et al., 2013; Petheram et al., 2012). Physical proximity methods choose the most similar gauged basin based on characteristics, addressing local variations. Conversely, these require detailed data and careful assessment of basin similarities (Di Prinzio et al., 2011). Regression-based methods link calibrated parameters to basin characteristics, enabling parameter estimation for ungauged basins without the need for selecting specific donor basins (Wagener and Wheater, 2006). Nevertheless, this requires a dataset of gauged basins large and diverse enough to establish statistically robust relationships between catchment attributes and model parameters for accurate implementation. A limited or geographically homogeneous data set may fail to capture the heterogeneity of hydrological responses, leading to poor generalization. Although both physically based and conceptual models are valuable for predicting river discharge under diverse runoff scenarios, they often demand extensive computational resources and are susceptible to uncertainties in input data and model structure, particularly when applied to ungauged basins. While multi-model ensembles (Pavan and Doblas-Reyes, 2000) and model calibration methods (Liu and Gupta, 2007; Neuman, 2003) have been explored to reduce bias, achieving reliable performance in ungauged conditions remains a challenge.

In contrast, machine learning models rely entirely on historical data to learn temporal patterns. They excel at capturing complex, nonlinear temporal dynamics and adapting to newly acquired information (Espeholt et al., 2022; Kratzert et al., 2018). However, these models face limitations, including a lack of interpretability, the need for large datasets, and sensitivity to hyperparameter selection. Although fewer machine learning models have been developed for predictions in ungauged basins compared to hydrological models, earlier approaches, such as Generalized Regression Neural Network (GRNN) (Besaw et al., 2010), and Counter Propagation Network (CPN) (Besaw et al., 2010), showed initial promise. More recently, deep learning architectures, particularly Long Short-Term Memory (LSTM) networks (Hochreiter and Schmidhuber, 1997; Kratzert et al., 2019a), have emerged as the state-of-the-art, demonstrating superior capability in capturing complex temporal dynamics. When applied to ungauged basins, these models typically incorporate basin characteristics as input, training neural networks on data from gauged basins and applying the trained models to ungauged ones. These machine-learning models, specifically LSTM, have been shown to outperform traditional hydrological models in terms of prediction accuracy in both gauged and ungauged settings (Kratzert et al., 2018, 2019a). However, training LSTMs, as well as architectures like GRNN and CPN, can be computationally expensive due to their complex architectures and often necessitate retraining when applied to different systems. In addition, many of these studies are validated in relatively data-rich scenarios where the majority of basins are treated as "gauged" during cross-validation, and only a limited number are treated as "ungauged." Consequently, these models benefit from abundant training data, which does not reflect the conditions in truly data-scarce regions, thereby limiting the relevance of such evaluations for ungauged basin prediction. For instance, the original LSTM model has demonstrated high predictive accuracy in ungauged basins through 12-fold cross-validation, but because the majority of basins in this setup are used for training, its predictive capability in genuinely data-scarce regions remains unproven (Kratzert et al., 2018). Furthermore, these models are often regarded as "black boxes," offering limited interpretability regarding how basin characteristics influence model predictions. To address this issue, a variant called Entity-Aware LSTM (EA-LSTM) has been developed to improve interpretability (Kratzert et al., 2019b). While EA-LSTM has been tested in ungauged settings, findings suggest that its ability to generalize across basins based on static features remains limited, with model performance largely relying on dynamic meteorological inputs (Heudorfer et al., 2025).

Although research on hybrid models that combine hydrological and machine learning approaches remains limited, these models offer promising potential for improving predictions in ungauged river basins (Tsai et al., 2021). By integrating the strengths of hydrological models, which provide physically consistent runoff estimations, with the ability of machine learning to capture complex, nonlinear relationships among variables, hybrid models can help mitigate structural biases inherent in traditional hydrological modeling (Cho and Kim, 2022; Lei et al., 2024; Yang et al., 2023; Zhou et al., 2022). This synergy presents a promising pathway for enhancing discharge prediction in ungauged basins.

Currently, modeling frameworks such as the Modular Assessment of Rainfall–Runoff Models Toolbox (MARRMoT) (Knoben, 2019), which includes 46 hydrological model structures, and the Raven hydrological modeling framework (Craig et al., 2020), which provides over 100 process algorithms and 80 interchangeable options for flexible model configurations, facilitate the implementation of multi-model ensembles. However, the integration of such multi-model ensembles into hybrid modeling frameworks has so far been limited.

In addition, a class of machine learning models known as reservoir computing (RC) (Jaeger and Haas, 2004) has gained attention in other scientific fields because of to its ability to learn and predict time series with low computational cost stemming from its non-iterative training (simple linear regression at the output layer) and high interpretability (Bianchi et al., 2021; Chattopadhyay et al., 2020; Sumi et al., 2023; Tomizawa and Sawada, 2021). Despite its potential, applications of RC in rainfall–runoff prediction remain scarce (De Vos, 2013). By combining RC with multi-model ensemble techniques, it may be possible to overcome key limitations of current hybrid models, achieving improved prediction accuracy, reduced computational demands, and enhanced interpretability. Such an approach could be particularly effective for discharge prediction in ungauged basins.

In this research, we propose a novel framework, **HY**drologic **P**rediction with multi-model **E**nsemble and **R**eservoir computing (HYPER). Instead of relying on parameter calibration, we posit that a sufficiently large ensemble of uncalibrated conceptual models can act as a broad prior distribution of hydrological behaviors. HYPER integrates this multi-model ensemble to reduce model bias and employs reservoir computing (RC) to achieve accurate, interpretable, and computationally efficient daily predictions in ungauged basins. Unlike event-based models that focus solely on extremes, HYPER reconstructs the continuous daily hydrograph from historical meteorological records, enabling the assessment of both flood peaks and persistent low-flow sequences essential for water supply management. We also aim to develop a method for transferring trained parameters from gauged to ungauged basins while preserving the interpretability of model components. First, we introduce a model architecture that combines multi-model ensembles with RC and evaluate its performance on gauged basins. Subsequently, we will test the framework on pseudo-ungauged basins, defined here as gauged basins where observed discharge is withheld during training to rigorously validate predictive accuracy by comparing different parameter estimation methods to identify the most accurate approach. By analyzing variations of HYPER, we will propose the optimal combination of multi-model ensemble and reservoir computing for predicting daily river discharge in ungauged basins.

## 2 Method

### 2.1 HYPER

#### 2.1.1 Hydrological models

We used the Modular Assessment of Rainfall-Runoff Models Toolbox (MARRMoT) v2.1. (Trotter et al., 2022), which provides a standardized framework for 47 distinct lumped conceptual hydrological models. The 47 models represent distinct hydrological models, such as HBV-96 (Burnash, 1995; Lindström et al., 1997), VIC (Liang et al., 1994), Sacramento Catchment Model (Burnash, 1995), and TOPMODEL (Beven, 1997). These models range in complexity from simple bucket models, such as “collie1” (ID:1) (1 parameter and 1 store), to complex structures like PRMS (ID:45) (18 parameters and 7 stores) and IHM19 (ID:47) (12 parameters, 8 stores), encompassing a diverse range of dominant hydrological process representation (e.g., saturation excess vs. infiltration excess mechanisms). The models utilize daily total precipitation, daily mean temperature, and daily mean potential evapotranspiration (PET) to estimate daily catchment-averaged runoff.

A key aspect of our framework is the use of uncalibrated models. While conceptual models are traditionally calibrated to individual basins to optimize performance, our objective was to evaluate the capability of the HYPER framework to function in ungauged basins where calibration data is unavailable. To achieve this, we employed the default parameter ranges provided by MARRMoT, which are standardized from a wide review of literature to be applicable across diverse hydro-climatic conditions. For each model, the initial parameter set was defined as the arithmetic mean of the feasible parameter range. In this context, the 47 uncalibrated models function as a broad prior distribution of potential hydrological behaviors. By including a structurally diverse ensemble, which spans various combinations of storage configurations and flux mechanisms, we aim to capture the uncertainty space of the catchment response without reliance on site-specific calibration. Although individual uncalibrated models may exhibit biases or fail to represent specific local processes, such as complex groundwater interactions, the ensemble approach is intended to span the true discharge dynamics, allowing subsequent post-processing to identify and weight the most appropriate structures.

### 2.1.2 Bayesian Model Averaging (BMA)

Previous studies demonstrate that applying BMA over all available models provides a higher prediction accuracy compared to selecting the single best model (Madigan and Raftery, 1994) or using simple arithmetic averaging (Tian et al., 2012). Therefore, we employed BMA to synthesize the outputs of the multiple hydrological models. BMA is a statistical method that combines predictions by assigning a weight to each model based on its relative performance. These weights reflect the likelihood of each model being the most appropriate given the observed data. Using these weights, we can calculate the weighted average of the ensemble (Hoeting et al., 1999).

To assess the performance of each model, the model outputs are compared to observations, and weights representing the fit to the observed value are calculated, typically through likelihoods. For the target variable $Q$, in our study, the river discharge, we calculate the posterior distribution as Eq. ( 1 ) where $M = \{M_1, M_2, M_3, \dots, M_n\}$ are the models considered, and D is the observed dataset.

$$P(\mathrm{Q}|D) = \sum_{k=1}^{K} P(\mathrm{Q}|M_k, D) P(M_k|D) \tag{1}$$

This posterior distribution of $Q$ is the average of the posterior distributions $\mathrm{P(Q|M_k, D)}$ for all models, with each model weighted by its corresponding posterior model probability $\mathrm{P(M_k|D)}$, hereafter referred to as the BMA weights. The BMA weight for a given model $M_k$ can be expressed as Eq. (2), where the likelihood of model $M_k$ being the true model is calculated as Eq. (3). Here, $\theta$ represents the vector of model parameters for model $M_k$. We assume a uniform prior probability for all models. While this assigns equal initial weight to every ensemble member, the Bayesian update process ensures that models exhibiting high predictive skill are assigned higher posterior weights, while poor-performing or redundant models are assigned lower posterior weights based on their likelihood scores.

$$P(M_k|D) = \frac{P(D|M_k)P(M_k)}{\sum_{l=1}^{k} P(D|M_l)P(M_l)} \tag{2}$$

$$P(D|M_k) = P(D|\theta, M_k) \tag{3}$$

Since streamflow distributions typically do not follow a Gaussian distribution due to heteroscedasticity and skewness, previous studies have developed methods to address this, such as using the Box-Cox transformation to stabilize variance and normalize residuals (Duan et al., 2007) or employing Copula-BMA to model non-linear dependencies without assuming a specific marginal distribution (Madadgar and Moradkhani, 2014). Consequently, there is often no theoretical requirement for model outputs to strictly follow a Gaussian distribution.

Although we tested both of these methods, we found that using BMA under the simple assumption that observational errors follow a Gaussian distribution achieved the highest accuracy for our specific ensemble of uncalibrated models. The underlying reason is twofold. First, the use of Box-Cox transformations can distort the error magnitude in the transformed space, potentially de-emphasizing the penalties for peak flow errors when back-transformed. Second, the use of Copula-BMA tends to restrict the ensemble output to the range defined by the individual models. Since we utilize uncalibrated conceptual models, individual members frequently exhibit significant biases, with some members overestimating and others underestimating the observations. A Gaussian BMA allows these positive and negative biases to cancel out linearly. In contrast, Copula-based methods, by preserving the dependence structure of the inputs, often fail to capture the true discharge when it lies outside the envelope of the uncalibrated ensemble, leading to a decrease in accuracy.

Therefore, to evaluate the likelihood $P(D|M_k)$, we used the log-likelihood under the assumption that the observational errors follow a Gaussian distribution centered at the model prediction. Specifically, the log-likelihood is computed based on the squared errors between observed and predicted values, as this formulation effectively prioritizes the reduction of mean squared error across the full hydrograph.

### 2.1.3 Reservoir Computing (RC)

Reservoir computing (RC) is a type of machine learning technique that utilizes an artificial recurrent neural network (RNN) to learn nonlinear systems. RC consists of an input layer, a reservoir, and an output layer, and is unique in that only the output layer is trained. Therefore, the fixed nature of the input layer and nodes of the reservoir allows RC to be trained solely on linear regression, rather than an iterative optimization process such as those used for gradient descent, making it computationally efficient. Once the parameters are trained for a certain system, there is no need to retrain the parameters (Pathak et al., 2018). Furthermore, unlike deep neural networks, where feature influence is obscured by multiple hidden layers, the RC output layer allows for a direct interpretation of how the reservoir's high-dimensional states map to the target discharge, a property we later exploit for regionalized predictions in ungauged basins.

The schematic of RC is illustrated in Fig. 1. The input data, the meteorological data, are an $M$-dimension time series, and the vector input for timestep $t$ is presented as $\boldsymbol{i}(t)$. The training data, the observed river discharge, is a timeseries and for timestep $t$ is presented as $\boldsymbol{u}(t)$. The RC's input layer and reservoir use randomly selected weights to capture the hidden states. The input layer uses $\boldsymbol{W}_{in}$, a $D \times M$-dimensional matrix to project the input vector to the reservoir space following Eq. (4).

$$\widehat{\boldsymbol{R}}_{in}[\boldsymbol{u}(t)] = \boldsymbol{W}_{in}\boldsymbol{i}(t) \tag{4}$$

The reservoir state at timestep $t$ can be expressed using a $D$-dimensional vector $\boldsymbol{r}(t)$. The reservoir state evolves per timestep based on Eq. (5), where $\boldsymbol{A}$ is the $D \times D$ adjacency matrix which represents the weighted and directed network of the $D$ nonlinear neurons.

$$\boldsymbol{r}(t + \Delta t) = \tanh\left[\boldsymbol{A}\boldsymbol{r}(t) + \boldsymbol{W}_{in}\boldsymbol{i}(t)\right] \tag{5}$$

The projection of the output layer can be formed as Eq. (6) to calculate the output vector, hereafter called RC weights, where $\boldsymbol{W}_{out}$ is a $D \times M$ matrix and $\boldsymbol{r}^{*}(t)$ represents a nonlinear transformation of $\boldsymbol{r}(t)$. Following the methods used by Chattopadhyay et al. (2020) and Tomizawa & Sawada (2021), the nonlinear transformation is based on Eq. (7) where $r_j$ is the $j$th element of $\boldsymbol{r}(t)$.

$$\widehat{\boldsymbol{R}}_{out} = \boldsymbol{W}_{out}\boldsymbol{r}^{*}(t) \tag{6}$$

$$r_j^{*} = \begin{cases} r_j & (\, j \text{ is even}) \\ r_{j-1} \times r_{j-2} & (\, j \text{ is odd}) \end{cases} \tag{7}$$

When we assume we have training data $\boldsymbol{u}(t)$ for $-T \leq t \leq 0$, the RC can train the output layer during this duration and compute the prediction of $\boldsymbol{u}(t)$, which we will denote as $\widetilde{\boldsymbol{u}}(t)$ during $t > 0$. During the training phase, the switch shown in Figure 1 lies horizontally and retrieves the input data and the observed training data. We use Tikhonov regularized linear regression to minimize the error between the training data and the prediction $\widetilde{\boldsymbol{u}}(t) = \boldsymbol{W}_{out}\boldsymbol{r}^{*}(t)$. To do so, we minimize the value computed using Eq. (8) where $\|X\| = X^{T}X$ and $\beta$ is the ridge regression parameter, set as a small positive value. Although this loss function assumes independent errors, the temporal correlations and autocorrelations inherent in discharge data are explicitly encoded within the evolving reservoir state $r(t)$ (Lukoševičius and Jaeger, 2009). Thus, the reservoir's dynamic memory allows the linear readout to capture complex temporal dependencies using a standard regression loss.

$$\sum_{n=1}^{\frac{T}{\Delta t}} ||\boldsymbol{u}(-n\Delta t) - \widetilde{\boldsymbol{u}}(-n\Delta t)||^2 + \beta||\boldsymbol{W}_{out}||^2 \tag{8}$$

During the prediction phase, the switch shown in Figure 1 is set vertically, so that only the input data is retrieved. We use the trained RC weights $\boldsymbol{W}_{out}$ to compute the prediction Eq. (9) based on the reservoir state Eq. (10).

$$\widetilde{\boldsymbol{u}}(t + \Delta t) = \boldsymbol{W}_{out}\boldsymbol{r}^{*}(t + \Delta t) \tag{9}$$

$$\boldsymbol{r}(t + \Delta \mathrm{t}) = \tanh\left[\boldsymbol{A}\boldsymbol{r}(t) + \boldsymbol{W}_{in}\boldsymbol{i}(t)\right] \tag{10}$$

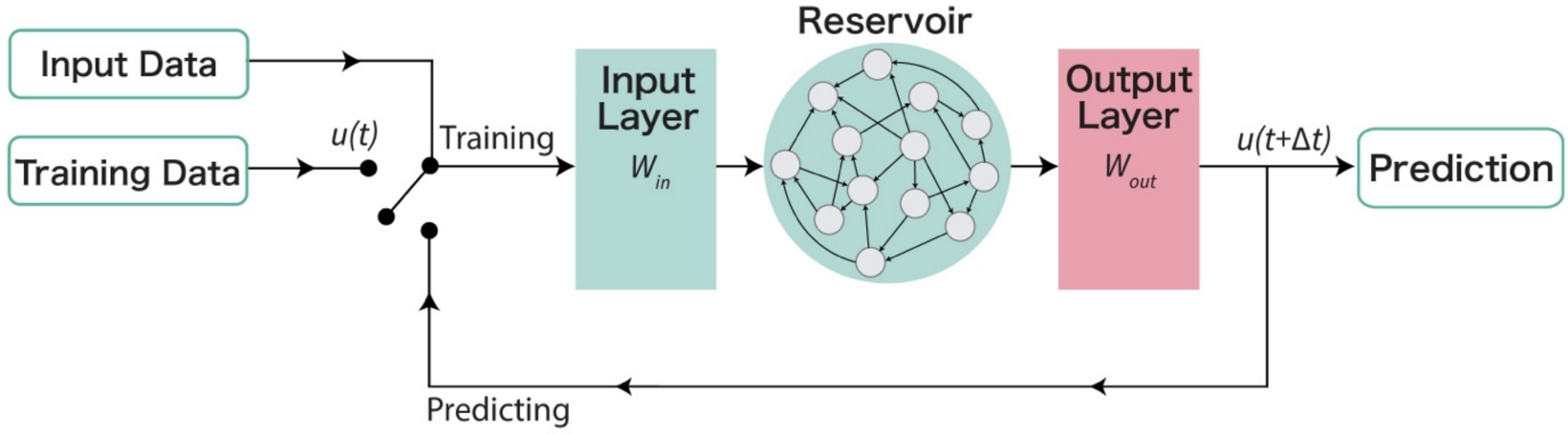

Figure 1 Schematic of reservoir computing (RC).

### 2.1.4 Bias Correcting Reservoir Computing (HYPER-BC)

Bias Correcting Reservoir Computing (HYPER-BC) (Fig. 2) uses the original RC framework to correct the bias of the conceptual hydrological models. The inputs are the daily total precipitation, daily mean temperature, and daily mean PET. We defined the target training data as the residual error between the observed daily basin-averaged discharge and the multi-model ensemble prediction. The RC is trained to predict this model error, and the predicted error is subsequently added to the ensemble prediction to compute the outflow during the prediction phase.

The prediction output of the multi-model ensemble at timestep $t$ can be expressed as $\boldsymbol{K}(t)$. The ensemble model error $\boldsymbol{b}(t)$ is assumed as Eq. (11), where $\boldsymbol{u}(t)$ is the observed discharge. This error $\boldsymbol{b}(t)$ serves as the target variable for the RC training phase.

$$\boldsymbol{b}(t) = \boldsymbol{u}(t) - \boldsymbol{K}(t) \tag{11}$$

Using the meteorological input data $\boldsymbol{i}(t)$ and the error target $\boldsymbol{b}(t)$ as training data, the RC is trained following the procedure described in Sect. 0 to predict the ensemble bias, denoted as $\widetilde{\boldsymbol{b}}(t)$. The final bias-corrected discharge prediction $\widetilde{\boldsymbol{u}}_B(t)$ is then computed as the sum of the ensemble prediction and the predicted bias, calculated as Eq. (12).

$$\widetilde{\boldsymbol{u}}_{\boldsymbol{B}}(t) = \boldsymbol{K}[\boldsymbol{i}(t - \Delta t)] + \widetilde{\boldsymbol{b}}(t) \tag{12}$$

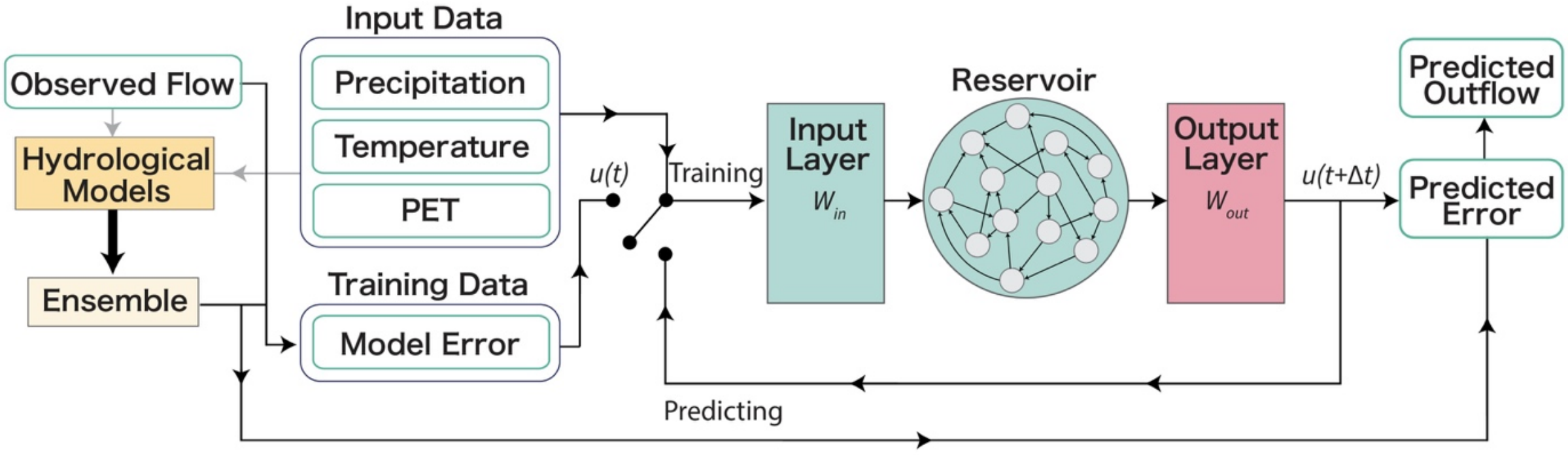

Figure 2 Schematic of Bias Correcting Reservoir Computing (HYPER-BC)

In addition to HYPER-BC, we investigated an alternative hybrid architecture referred to as Hybrid Reservoir Computing (HYPER-RCH). Based on the framework proposed by Pathak et al. (2018), HYPER-RCH integrates the physical model, in this case hydrological models, directly into the reservoir dynamics rather than as a post-processor. A detailed description of the HYPER-RCH framework is provided in the Supplementary Materials.

We included this architecture to rigorously evaluate whether bias correction or dynamic integration yields superior performance for hydrological applications. As presented in the Results section (Sect. 5.2), we conducted a comparative evaluation of both architectures using data from gauged basins. This analysis demonstrated that HYPER-BC consistently achieved higher accuracy than HYPER-RCH. Consequently, based on this empirical selection of the optimal architecture, the subsequent evaluation in this paper regarding ungauged basins focuses exclusively on the HYPER-BC framework.

**2.1.5 Predictions in Ungauged Basins with HYPER**

Once the BMA and RC weights are trained in gauged river basins (noting that our proposed method requires no calibration of hydrological models' parameters), the challenge is to effectively use these weights to infer the appropriate weights of ungauged river basins. In this study, we evaluated two regionalization methods to estimate BMA and RC weights for ungauged basins: spatial proximity and regression-based inference. Note that all hydrological modeling and RC training in this study were conducted using specific discharge (mm/day). This normalization ensures that the learned weights reflect hydrological process dynamics per unit area, rather than being dominated by the magnitude of differences of volumetric flow.

The process of applying the spatial proximity method is illustrated in Fig. 3 (a). Based on the Euclidean distance between the centroids of the gauged and ungauged basins, the closest gauged basin is selected as the donor. The ungauged basin adopts the BMA weights and RC weights of the donor basin.

The process of applying regression-based inference is shown in Fig. 3 (b). The aim is to learn a mapping between observable basin characteristics (e.g., topography, geology, climate indices, land use, soil cover) and the optimized BMA and RC weights. This approach is based on the premise that physical catchment attributes dictate the dominant runoff generation mechanisms and the memory of the system. Therefore, these attributes should conceptually predict which model structures are most suitable and the nature of residual error dynamics.

Using this learnt relationship, we can estimate the weights for ungauged basins. To do so, we concatenate the 47 BMA weights and the $D$-dimensional RC output layer weights into a single weight vector. For the gauged basins, we standardize the vector, as BMA weights are limited to a range of $[0, 1]$ whereas RC output layer weights do not have a bounded range.

We apply Principal Component Analysis (PCA) (Uyeda et al., 2015) to reduce the dimension of the weight vector. The PCA transformation matrix is derived exclusively from the gauged basin set to prevent information leakage. Let $s$ be the number of gauged basins and $t$ be the number of weight parameters ($t = 47 + D$). The $m \times m$ covariance matrix $\boldsymbol{R}$ is computed by Eq. (13) where $m = \min(s, t)$, $\boldsymbol{Z}$ is the standardized dataset with dimension $s \times t$ containing all basin weight parameters, $\boldsymbol{\mu}$ is the mean of the $t$ parameters, and $\mathbf{1}$ is a column vector of ones. The principal component score matrix $\boldsymbol{S}$ can be calculated as Eq. (14), where the columns of $\boldsymbol{V}$ are the eigenvectors of $\boldsymbol{R}$.

$$\boldsymbol{R} = (s-1)^{-1}(\boldsymbol{Z} - \boldsymbol{1\mu^T})^T(\boldsymbol{Z} - \boldsymbol{1\mu^T}) \tag{13}$$

$$\boldsymbol{S} = (\boldsymbol{Z} - \boldsymbol{1\mu^T})\boldsymbol{V} \tag{14}$$

Using the principal components, we apply Lasso regression to define the relationship with the standardized basin attributes. Lasso regression is a linear regression where the coefficients can be set to zero, allowing for a more interpretable result (Tibshirani, 1996). Regarding the non-linearity of basin area, note that the regression target here is the latent model weight vector, not the discharge magnitude itself. Therefore, the relative contribution of the weights is hypothesized to scale more linearly to the basin attributes.

When $p$ principal components and n basin attributes are considered, the vector $\boldsymbol{y}$ containing the $p$ principal component values and the $p \times n$ matrix $\boldsymbol{X}$ containing basin attributes can be modeled as Eq. (15). Here, vector $\boldsymbol{\beta}$ contains $n$ values, representing the unknown regression coefficients, and vector $\boldsymbol{\varepsilon}$ contains the error. The coefficients are estimated based on Eq. (16) where $\lambda$ is the tuning parameter.

$$\boldsymbol{y} = \boldsymbol{X\beta} + \boldsymbol{\varepsilon} \tag{15}$$

$$\min_{\boldsymbol{\beta}}(\boldsymbol{y} - \boldsymbol{X\beta})^T(\boldsymbol{y} - \boldsymbol{X\beta}) + \lambda \sum_{i=1}^{n} |\beta_i|, \qquad \lambda \geq 0 \tag{16}$$

Once the regression coefficients are estimated for the gauged basins, we apply these coefficients to the ungauged basins to compute their principal components from their basin attributes. Then, we apply the eigenvectors computed for the gauged basins to invert the PCA process following Eq. (17) to compute the standardized ungauged basin dataset $\boldsymbol{Z_R}$, containing the estimated weight parameters in the original vector dimension. Finally, we invert the standardization process using the mean and standard deviation of the gauged basin dataset to recover the final weight parameters for the ungauged basins.

$$\boldsymbol{Z_R} = \boldsymbol{ZV} \tag{17}$$

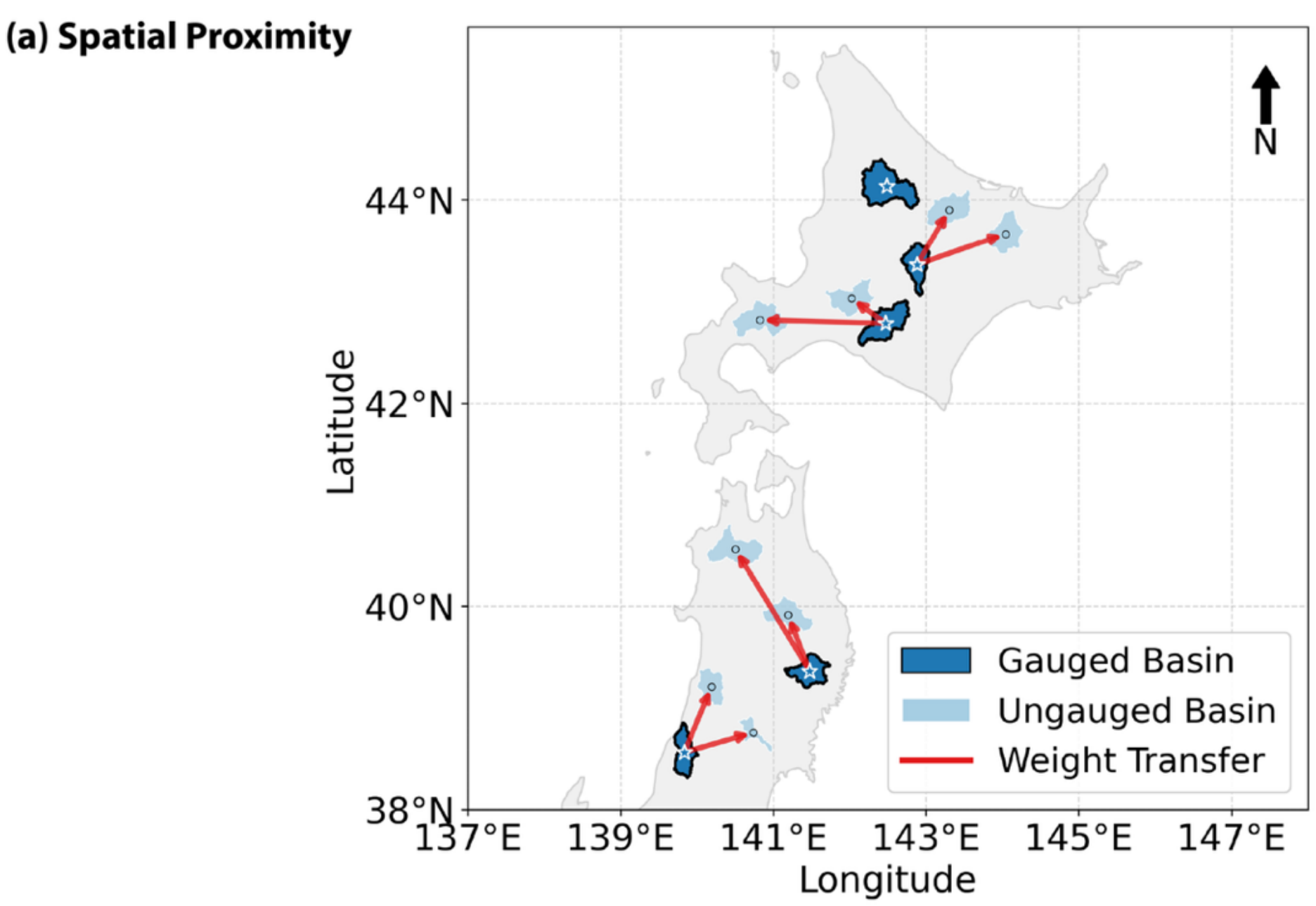

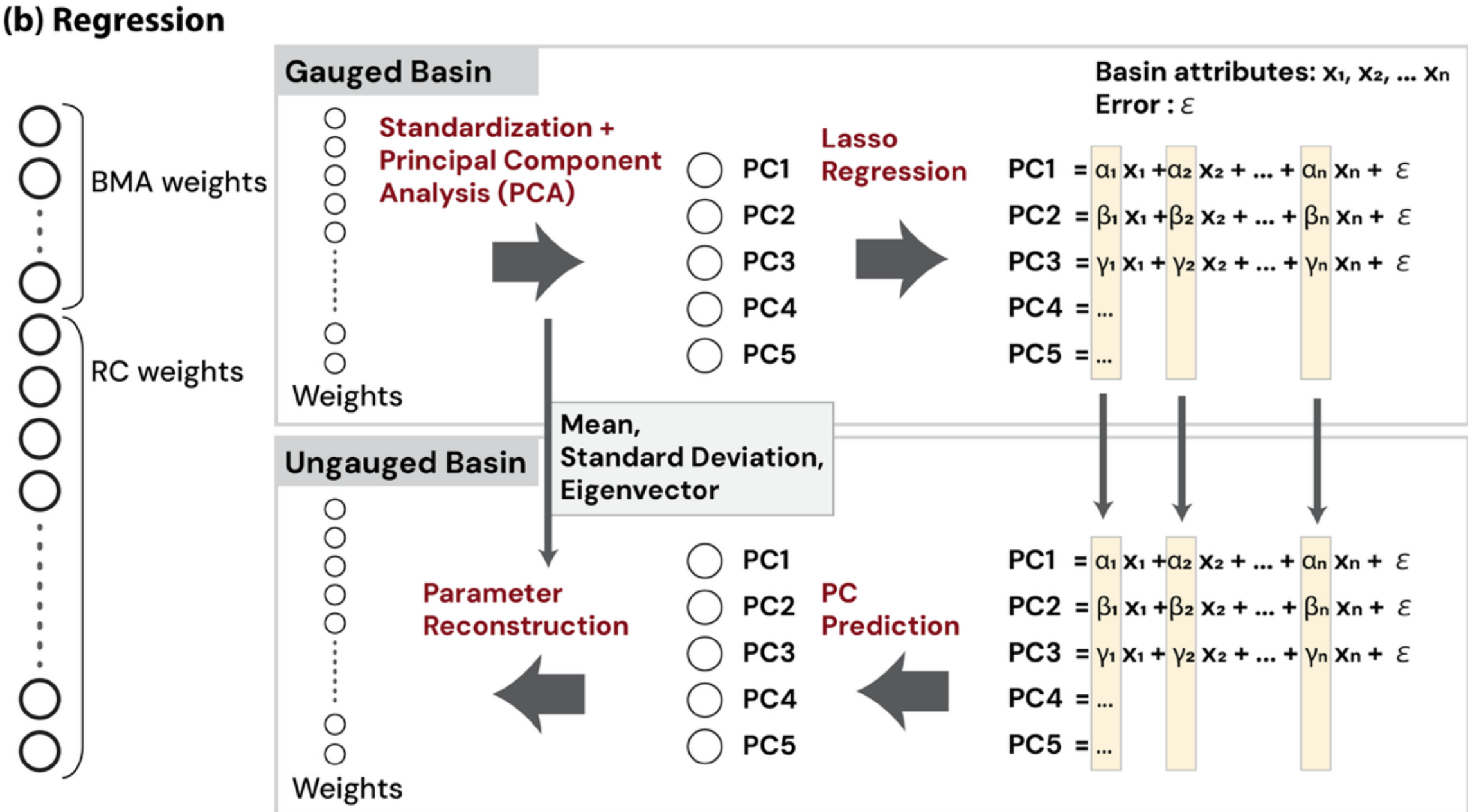

Figure 3 Diagram of (a) spatial proximity and (b) regression-based methods for prediction in ungauged basins.

### 2.2 Long-Short Term Memory (LSTM)

As a benchmark against the HYPER-BC model, we use the Long Short-Term Memory (LSTM) model. LSTM is a deep learning architecture capable of learning long-term dependencies in sequential data. It has been developed and evaluated on both gauged and ungauged basins, showing comparable or higher prediction accuracy compared to traditional hydrological models, demonstrating state-of-the-art performance in rainfall-runoff modeling (Kratzert et al., 2018, 2019a).

LSTM is a variant of RNN characterized by a memory cell regulated by three gating mechanisms: the forget gate, input gate, and output gate, along with two states: the hidden state and the cell state. These gates modulate the flow of information, allowing the network to retain relevant long-term hydrological states while discarding irrelevant data. The forget gate determines what information to forget and remove from the memory cell, the input gate determines what new information to add to the memory cell, and the output gate controls what information is output from the memory cell. The hidden state is recomputed at each timestep, while the cell state stores long-term information.

LSTM takes the sequential meteorological forcing data from the preceding T timesteps ($1 \leq t \leq T$) as input to predict the discharge at $t = T + 1$. During the training, the model learns by adjusting its parameters to minimize the loss between the observed data and predicted outputs. Once trained, the model uses these learnt parameters to generate predictions for each subsequent timestep.

A critical distinction in this study is the training strategy employed for the benchmark versus the proposed framework. While HYPER operates on a basin-specific basis (training local weights and transferring them), the LSTM utilizes a regional training strategy (Kratzert et al., 2019a). In the regional approach, a single LSTM model is trained simultaneously on concatenated data from all available basins. This allows the model to learn a universal rainfall-runoff mapping that generalizes across diverse hydro-climatic conditions, effectively treating the entire dataset as one large basin.  This global data assimilation is a primary strength of the LSTM, often yielding higher accuracy than basin-by-basin training. By adopting this regional strategy for the benchmark, we subject the basin-specific HYPER framework to the most rigorous "upper-bound" to test whether a computationally efficient, local-transfer approach can compete with a globally optimized deep learning model.

### 2.2.1 Prediction in Gauged Basins

For gauged basins, we used the standard LSTM configuration (Kratzert et al., 2018). It was trained on the regional dataset using daily precipitation, temperature, and PET, identical to those used in the HYPER model. During training, the observed daily basin-averaged river discharge is used as the target, while during prediction, the model simulates the daily basin-averaged river discharge. The mean squared error (MSE), defined in Eq. (18), was used as the loss function, where $n$ is the number of data points, $obs_i$ is the observed values, and $sim_i$ is the simulated values.

$$\mathrm{MSE} = \frac{1}{n}\sum_{i=1}^{n}(obs_i - sim_i)^2 \tag{18}$$

### 2.2.2 Prediction in Ungauged Basins

For ungauged basins, we adopted the Entity-Aware LSTM model developed by Kratzert et al. (2019). Static basin attributes were concatenated with the meteorological time-series inputs at each time step. This enables the LSTM to learn a mapping not just from weather to flow, but from basin characteristics to flow dynamics, allowing it to generalize to ungauged basins. To ensure a fair comparison with HYPER, the LSTM was optimized using MSE rather than the NSE-based loss often used in hydrological applications.

## 2.3 Evaluation Metrics

To provide a comprehensive assessment of model performance, we selected four metrics that evaluate distinct statistical properties of the hydrograph, which are Nash-Sutcliffe Efficiency (NSE) (Nash and Sutcliffe, 1970), Kling-Gupta Efficiency (KGE) (Gupta et al., 2009), E1 (Legates and McCabe Jr., 1999), and Root Mean Square Error (RMSE).

Nash-Sutcliffe Efficiency (NSE) is based on the squared differences (L2-norm) between observed and simulated values, as shown as Eq. (19). Because it squares the residuals, NSE is highly sensitive to large errors, making it a primary metric for evaluating peak flow accuracy. It measures how well the predicted values match the observed data, and the values lie within the range $(-\infty, 1]$. A value closer to 1 indicates better model performance, while values closer to 0 or negative values suggest poor model performance (see Duc and Sawada, 2023).

$$NSE = 1 - \frac{\sum_{i=1}^{n}(obs_i - sim_i)^2}{\sum_{i=1}^{n}\left(obs_i - \overline{obs}\right)^2} \tag{19}$$

Kling-Gupta Efficiency (KGE) addresses the limitations of single-objective metrics by decomposing error into three distinct components, correlation, bias ratio, and variability ratio, calculated as Eq. (20), where $r$ is the Pearson correlation coefficient. This provides a balanced view of model performance, ensuring that a high score reflects not just goodness-of-fit, but also the correct reproduction of flow variance and mean volume. KGE values lie within the range $(-\infty, 1]$ with values closer to 1 indicating better performance.

$$KGE = 1 - \sqrt{(r-1)^2 + \left(\frac{\mu_{sim}}{\mu_{obs}} - 1\right)^2 + \left(\frac{\sigma_{sim}}{\sigma_{obs}} - 1\right)^2} \tag{20}$$

Modified Efficiency (E1) utilizes the absolute differences (L1-norm) rather than the squared differences used for NSE. It is calculated as Eq. (21). By avoiding the squaring operation, E1 reduces sensitivity to outliers and extreme peaks, providing a more robust assessment of low-to-medium flow performance and overall water balance. Values lie within the range $(-\infty, 1]$ with a value closer to 1 indicating better model performance.

$$E1 = 1 - \frac{\sum_{i=1}^{n}|obs_i - sim_i|}{\sum_{i=1}^{n}|obs_i - \overline{obs}|} \tag{21}$$

Root Mean Square Error (RMSE) is calculated as Eq. (22) and quantifies the magnitude of prediction errors by taking the square root of the average squared differences between observed and predicted values. Values lie within the range $[0, \infty)$ with values closer to 0 indicating lower error. While mathematically related to NSE, RMSE is an absolute metric, meaning it is inherently sensitive to flow magnitude and will be naturally larger for "flashy" basins. In contrast, NSE is a relative metric that normalizes performance by basin variance, placing all basins on a comparable scale. We include both to distinguish whether model errors are driven by flow magnitude (RMSE) or by structural deficiencies independent of scale (NSE).

$$RMSE = \sqrt{\frac{1}{n}\sum_{i=1}^{n}(obs_i - sim_i)^2} \tag{22}$$

Finally, we evaluated the computational cost of each framework. Specifically, we measured the computing time required for training and prediction using a single Intel Xeon Gold CPU processor. While model calibration is often viewed as a one-time cost, computational efficiency remains a constraint for global-scale applications involving millions of basins and for uncertainty quantification, which requires running large ensembles (e.g., hundreds of iterations). Furthermore, reducing computational demands is essential for democratizing access to advanced hydrological modeling in regions with limited high-performance computing infrastructure.

## 3 Data

We utilized the Multi-model Ensemble for Robust Verification of hydrological modeling in Japan (MERV-Jp) ver2.0 (Sawada et al., 2022; Sawada and Okugawa, 2023). This dataset provides daily observed discharge, daily total precipitation, daily mean temperature, and daily mean PET for 87 basins in Japan (Fig. 4) covering the period from 1986 to 2015. Although MERV-Jp includes calibrated simulation outputs for a subset of MARRMoT models, our study relies on uncalibrated conceptual models to simulate ungauged conditions. Therefore, we utilized only the meteorological forcing data from this dataset to drive the 47 models in the MARRMoT v2.1 framework. A complete list of the 47 hydrological models used, along with their complexity (number of parameters and stores), is provided in the Supplementary Materials. The 87 study basins represent diverse hydrological scales, with drainage areas ranging from $275\ km^2$ to $12{,}697\ km^2$ (median: $1{,}740\ km^2$).

To implement the regression-based regionalization for prediction in ungauged basins, we derived the catchment attributes using GIS data from the National Land Information download portal for Japan (Ministry of Land, Infrastructure, Transport and Tourism, 2025). We processed shapefiles representing topography, meteorology, soil properties, geology, landform, and land use to extract a set of 154 static basin attributes. These attributes, which serve as predictors for the regression model, are listed in the Supplementary Materials.

## 4 Experiment Design

To systematically evaluate the prediction performance of the proposed methods under varying hydrological data conditions, we conducted four experiments. Experiment 1 tests the model's performance in gauged basins. Experiments 2 and 3 evaluate performance in pseudo-ungauged basins, defined here as gauged basins where observed discharge is withheld during the training phase to simulate ungauged conditions, under data-rich and data-scarce conditions, respectively. Experiment 4 assesses prediction accuracy in “remote” scenarios where gauged basins are geographically distant from pseudo-ungauged ones.

Table 1: Summary of Experiments Conducted

| Experiment | Scenario Description | Section |
|---|---|---|
| 1 | Prediction in gauged basins | 4.1 |
| 2 | Prediction in data-rich ungauged basins | 4.2.1.1 |
| 3 | Prediction in data-scarce ungauged basins | 4.2.1.2 |
| 4 | Prediction in remote ungauged basins | 4.2.1.3 |

### 4.1 Experiment 1: Prediction in Gauged Basins

To evaluate the HYPER models (Sect. 2.1), we assessed river discharge prediction accuracy using 87 gauged basins in Japan. The models compared included arithmetic averaging of the 47 MARRMoT models (AVE), Bayesian model averaging of the 47 MARRMoT models (BMA), reservoir computing without hydrological models (RC), the proposed HYPER models (HYPER-BC and HYPER-RCH), and the LSTM benchmark model. Additionally, 47 uncalibrated individual MARRMoT models were included for comparison. This comprehensive comparison allows us to evaluate the performance of HYPER-BC and HYPER-RCH against individual uncalibrated models, ensemble approaches, standalone RC, and the calibrated LSTM benchmark. Computational cost was also compared relative to the LSTM model.

The parameters used for standalone RC, HYPER-RCH, and HYPER-BC were selected based on a preliminary sensitivity analysis (details provided in Supplementary Material). These included a reservoir size of 700, an adjacency density of 0.0006, a spectral radius of 0.4, an input scale of 0.5, and a ridge regression parameter of 0.001. For LSTM, the model parameters followed (Kratzert et al., 2019b), with a learning rate of 0.001, a window size of 270, a batch size of 256, a dropout rate of 0.4, and a hidden size of 256. Regarding the training duration, we conducted a preliminary analysis testing up to 100 epochs to ensure convergence, given the smaller dataset size (87 basins) compared to the original study. This analysis confirmed that 30 epochs, the value used in the reference study, was optimal for our dataset, achieving convergence while preventing overfitting.

The experiment used observed basin-averaged daily discharge as training data, with daily total precipitation, daily mean temperature, and daily mean PET as input variables. The training period spanned eight years, from January 1, 1993, to December 31, 2000, and the validation period covered six years, from January 1, 2001, to December 31, 2006. Model performance was evaluated using NSE, KGE, E1, and RMSE for prediction accuracy, along with computational time per basin to assess computational efficiency.

### 4.2 Prediction in Ungauged Basins

First, we tested two methods for predicting in ungauged basins: the spatial proximity method, hereafter called HYPER-BcProx (**HYPER**-**B**ias **C**orrecting RC with Spatial **Prox**imity), and the regression-based method, hereafter called HYPER-BcReg (**HYPER**-**B**ias **C**orrecting RC with **Reg**ression). These methods were evaluated within Japan by dividing basins into gauged "training" basins and pseudo-ungauged "testing" basins.

We initially evaluated prediction accuracy under data-rich scenarios using k-fold cross-validation, a commonly adopted approach in previous studies, where most basins are considered gauged, and a small portion is treated as pseudo-ungauged. Since dense observation networks are rare globally, we also tested more data-scarce scenarios to assess model robustness under limited data conditions. Additionally, to examine prediction performance when ungauged basins are geographically distant from gauged basins, we evaluated the models in a spatially separated "remote" scenario.

For the reservoir computing component of the HYPER model in these regionalization experiments, hyperparameter analysis determined a reservoir size of 200, a spectral radius of 0.4, and a ridge regression parameter of 1.0. For HYPER-BcReg, the alpha parameter was set to 0.1, and the number of principal components (PCs) retained in the PCA was three, also based on hyperparameter tuning (see Supplementary Materials).

For the LSTM model, we used the hyperparameters based on Kratzert et al. (2019a), with a learning rate of 0.001, a window size of 270, a hidden layer size of 256, and a dropout rate of 0.4. Regarding training duration, we performed a preliminary sensitivity analysis testing up to 50 epochs within the k-fold cross-validation framework (described in Sect. 4.2.1.1) to ensure convergence on the regional dataset. This analysis confirmed that 30 epochs were optimal. While optimal epoch size may vary with training sample size, we maintained this fixed stopping criterion across all scenarios to emulate realistic operational conditions where validation data for tuning is unavailable. Therefore, this setting was applied across all ungauged basin experiments.

#### 4.2.1.1 Experiment 2: Predictions in Data-Rich Scenarios

We conducted k-fold cross-validation, a widely used technique for assessing prediction accuracy in ungauged basins. In this approach, the 87 basins were divided into k groups; following Kratzert, Klotz, Herrnegger, et al. (2019), we used 12 folds. Each fold considers one group of basins as "ungauged", while the remaining basins are treated as "gauged." Prediction accuracy is then assessed on the ungauged basins. This process is repeated 12 times so that each group is evaluated once as ungauged. The overall prediction accuracy is based on the combined results from all ungauged basins across the folds.

The training period spanned 8 years, from January 1, 1993, to December 31, 2000, and the prediction period covered 6 years, from January 1, 2001, to December 31, 2006. Model performance was evaluated using NSE, KGE, E1, and RMSE as accuracy metrics, along with the average computational time per basin.

#### 4.2.1.2 Experiment 3: Predictions in Data-Scarce Scenarios

To evaluate prediction accuracy for "ungauged" basins depending on the number of "gauged" basins, we tested several variations in the number of "gauged" basins. We first selected 17 "ungauged" testing basins (shown with hatching in Fig. 4), chosen sequentially based on their longitude. This number was chosen to provide sufficient variability to assess model accuracy across different regions in Japan, while leaving basins available to be used as "gauged" basins. From the remaining 70 basins, we selected $n$ "gauged" basins, with $n$ taking the values of 3, 10, 20, 30, 50, and 70 basins.

For each value of $n$, we conducted 100 iterations, where in each iteration a different set of $n$ "gauged" basins was randomly selected. Model training and prediction were then conducted using this selection. This process allowed us to estimate the average prediction accuracy and its variability as a function of the number of available "gauged" basins. For each iteration and each of the 17 fixed "ungauged" test basins, we computed the mean prediction accuracy using relevant evaluation metrics. To quantify prediction uncertainty, we also calculated the 90th percentile reliability range for each $n$, based on the distribution of median metric values across iterations.

#### 4.2.1.3 Experiment 4: Predictions in Remote Scenarios

To assess how prediction accuracy is affected when "gauged" basins are geographically distant from "ungauged" basins, we tested a remote scenario by restricting "gauged" basins to a single region rather than allowing them to be spatially dispersed. We used the same 17 fixed "ungauged" basins as in Experiment 3 (shown in hatches in Fig. 4) and divided the remaining 70 basins into regions as shown in Fig. 4. For each experiment, one region was designated as "gauged" and used to train the HYPER model, while prediction accuracy was evaluated on the fixed 17 "ungauged" basins. This process was repeated for each region. The prediction accuracy was compared to that obtained when a similar number of "gauged" basins (15 gauged basins) were randomly selected from all basins, as performed in Experiment 3. The LSTM model was trained using the same setup of regional "gauged" and fixed "ungauged" basins. All other experimental settings followed those in Experiment 2 (Sect. 4.2.1.1).

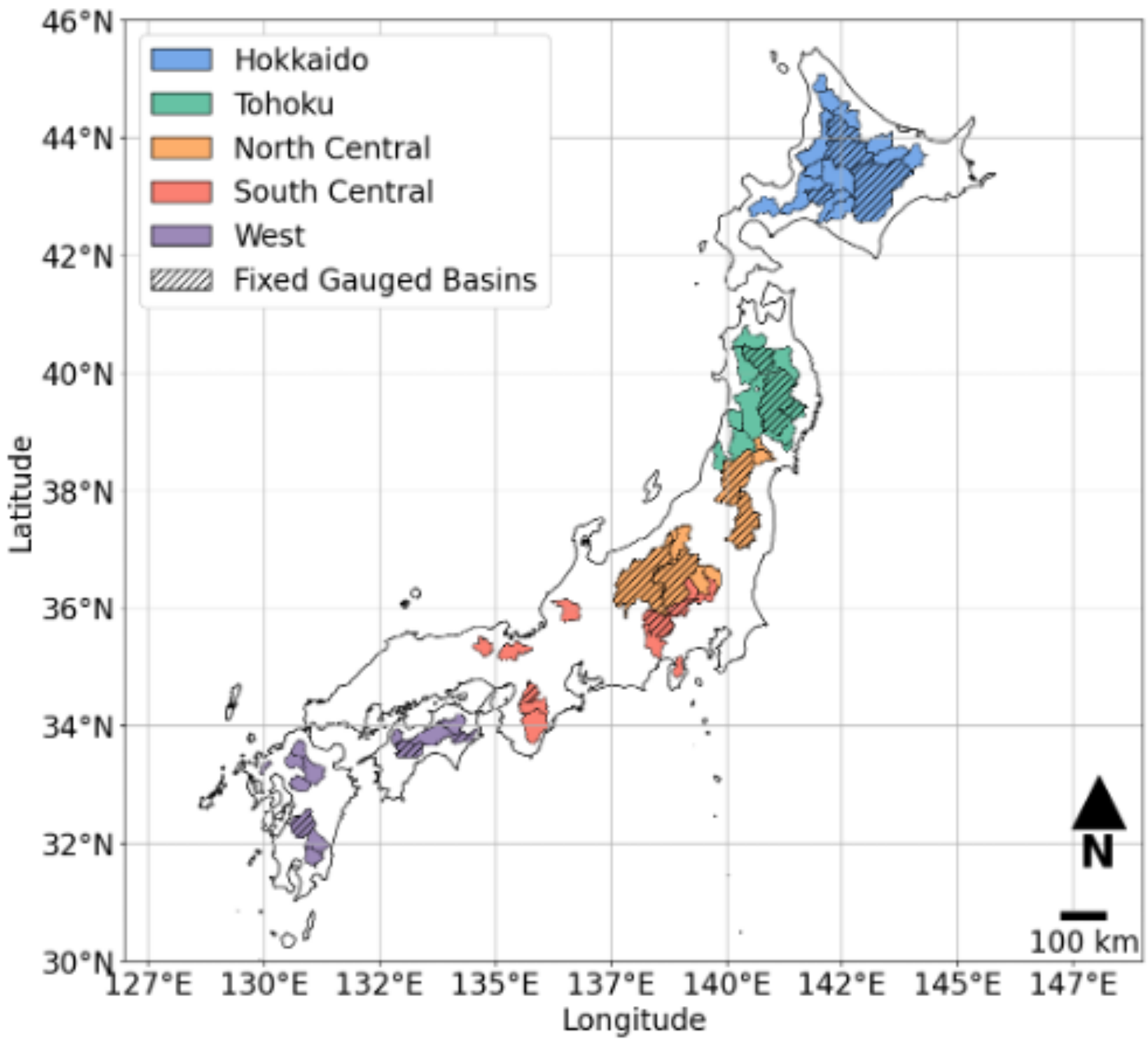

Figure 4. Study area. Hatched basins are treated as fixed 17 "ungauged" basins in Experiments 2 and 3.

## 5 Results

### 5.1 Experiment 1: Prediction in Gauged Basins

Figure 5 shows the performance of our hybrid models, HYPER-RCH and HYPER-BC, evaluated on gauged basins. For comparison, the results of individual uncalibrated MARRMoT models (shown in gray), the Bayesian Model Averaging (BMA) ensemble, the standalone Reservoir Computing (RC), and the LSTM model are also shown.

Both HYPER-RCH and HYPER-BC, which combine multi-model ensembles with Reservoir Computing, outperform all individual uncalibrated MARRMoT models as well as BMA and RC. Notably, a comparison with previous work providing the calibrated model outputs in these basins (Sawada and Okugawa, 2023) reveals that the uncalibrated BMA ensemble achieved higher accuracy and lower variability than the distribution of calibrated single models. This confirms that the aggregated diversity of the uncalibrated ensemble can surpass the performance of individual parameter-optimized models. These results highlight the advantage of integrating hydrological model diversity and data-driven methods to enhance predictive accuracy. Between the two hybrid approaches, HYPER-BC demonstrates the highest performance across most evaluation metrics.

When comparing HYPER-BC with the LSTM benchmark, we find that LSTM achieves higher predictive accuracy in terms of NSE, E1, and RMSE. The median values for LSTM are 0.77 for NSE, 0.55 for E1, and 2.12 for RMSE, while HYPER-BC shows median values of 0.62, 0.34, and 2.67 for the same metrics, respectively. This performance gap is expected, not only because LSTM is a deep learning model, but because squared-error metrics (like NSE and RMSE) disproportionately penalize errors in peak flows. Since the LSTM is optimized directly on Mean Squared Error (MSE), it effectively minimizes these peak errors, resulting in superior scores compared to HYPER-BC, which uses a closed-form linear regression approach. However, in terms of KGE, HYPER-BC performs comparably to LSTM, with both models achieving a median KGE of 0.75. These results suggest that while LSTM generally provides stronger performance on squared-error metrics, HYPER-BC still demonstrates competitive skill in capturing balanced runoff characteristics, including flow variability and bias.

The average computational time per basin for calibration and prediction using HYPER-RCH, HYPER-BC, and LSTM was evaluated. HYPER-RCH and HYPER-BC do not require iterative parameter calibration, while prediction takes approximately 3 minutes and 5 seconds per basin. The 3-minute component corresponds to a one-time forward simulation of the 47 MARRMoT models using MATLAB and its BMA ensemble. For LSTM, the training time is 6 minutes, and the prediction time is approximately 3 seconds per basin. Although HYPER-RCH and HYPER-BC eliminate the need for iterative calibration, their total runtime is comparable to that of LSTM due to the initial requirement to simulate the 47 MARRMoT models. However, since the MARRMoT simulations are performed only once and can be reused across multiple predictions without re-fitting, the incremental cost of prediction using HYPER-BC or HYPER-RCH becomes substantially lower in operational or large-scale applications. Additionally, because these methods do not involve a training process, they do not require recalibration when new data (e.g., an additional year of observations) becomes available. In contrast, LSTM models typically require retraining from scratch to incorporate new data, which adds to their computational cost and operational complexity.

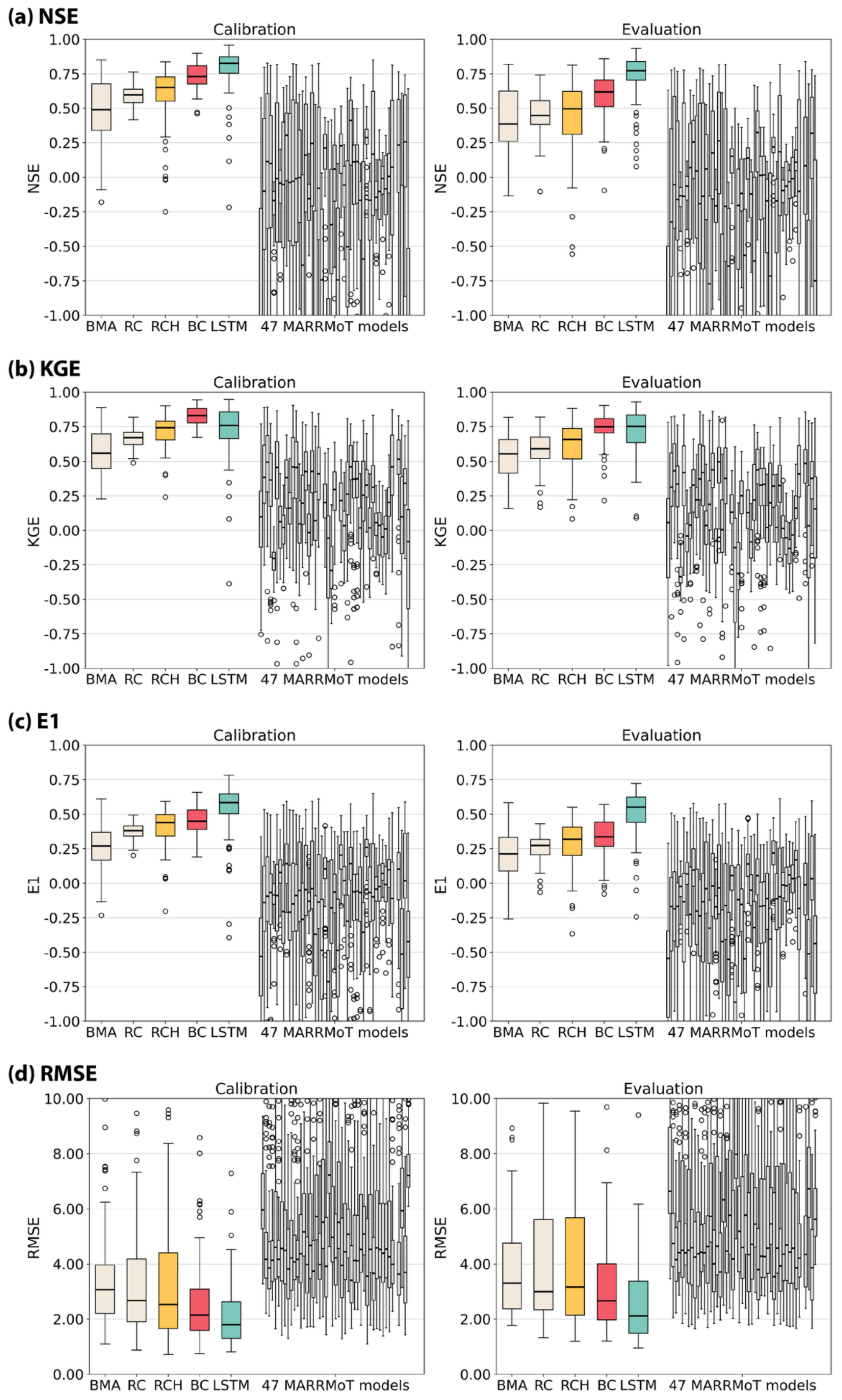

Figure 5 Model accuracy using (a) NSE, (b) KGE, (c) E1, and (d) RMSE benchmarks on 87 basins for the calibration and evaluation period. BMA (cream), RC (cream), HYPER-RCH (yellow), HYPER-BC (red), LSTM (green) and 47 MARRMoT models (gray) are shown.

## 5.2 Prediction in Ungauged Basin

### 5.2.1 Experiment 2: Predictions in Data-Rich Scenarios

Comparing HYPER-BcReg and HYPER-BcProx, we found that HYPER-BcReg has higher prediction accuracy (Fig. 6). A key strength of HYPER-BcReg lies in its interpretability, which we demonstrate by examining the Lasso regression coefficients for the first two principal components used in weight estimation (Fig. 7). The first principal component is primarily influenced by meteorological variables, particularly snow depth, precipitation, and sunshine duration, along with soil attributes like the percentage of gleysols. The second principal component is mainly shaped by topographical attributes such as mean slope, along with other properties related to meteorology, land use, and landform. Physically, the regression leverages these attributes to adapt the ensemble structure. For example, the first principal component, driven by snow depth and winter precipitation, effectively up-weights model structures designed for snowmelt and saturation-excess processes. This allows the framework to automatically shift predictive reliance toward physically appropriate models in northern, snow-dominated basins. This analysis confirms that meteorological and topographical factors are essential predictors in weight estimation. Furthermore, the use of Lasso regression promotes interpretability by allowing direct identification of the key catchment features influencing model weights, a significant advantage over purely data-driven black-box models.

Figure 6 compares prediction accuracy for HYPER-BcProx, HYPER-BcReg, and the benchmark LSTM model using a 12-fold cross-validation. For NSE, E1, and RMSE, LSTM achieves the highest accuracy with median values of 0.64 (NSE), 0.31 (E1), and 2.84 (RMSE), outperforming HYPER-BcReg, which has median values of 0.59, 0.33, and 2.85, respectively. This advantage likely stems from LSTM's ability to capture complex nonlinear temporal dynamics through deep learning and parameter optimization, whereas HYPER-BcReg uses a simpler linear regression approach. However, for the KGE metric, HYPER-BcReg outperforms the benchmark, reaching a median KGE of 0.65 compared to LSTM's 0.55. This indicates that while LSTM minimizes squared error more effectively, HYPER-BcReg provides a superior balance of correlation, bias, and variability.

Table 2 reports the computational time per basin for each model. The HYPER models require no iterative calibration, resulting in a markedly lower computational time per basin, approximately 3 minutes, compared to LSTM, which requires approximately 95 minutes due to its calibration process. Although HYPER models rely on the predictions of 47 MARRMoT hydrological models, these predictions are performed once and reused, ensuring that the incremental cost of prediction remains competitive (approx. 5 seconds per basin). This computational efficiency makes HYPER-BcReg suitable for large-scale or operational applications where speed and scalability are critical.

Overall, these results demonstrate that regression-based HYPER approaches provide fast, accurate, and interpretable modeling in data-rich settings, achieving performance close to deep learning methods in squared-error metrics while surpassing them in KGE, all with greater efficiency and transparency. It is important to note that in this 12-fold cross-validation, approximately 91.2% of basins are considered "gauged," reflecting a data-rich environment that may be rare in many global regions. Thus, evaluating model robustness in more realistic, data-scarce scenarios is vital. The subsequent experiment (Sect. 4.2.1.2) examines HYPER and LSTM performance when fewer "gauged" basins are available.

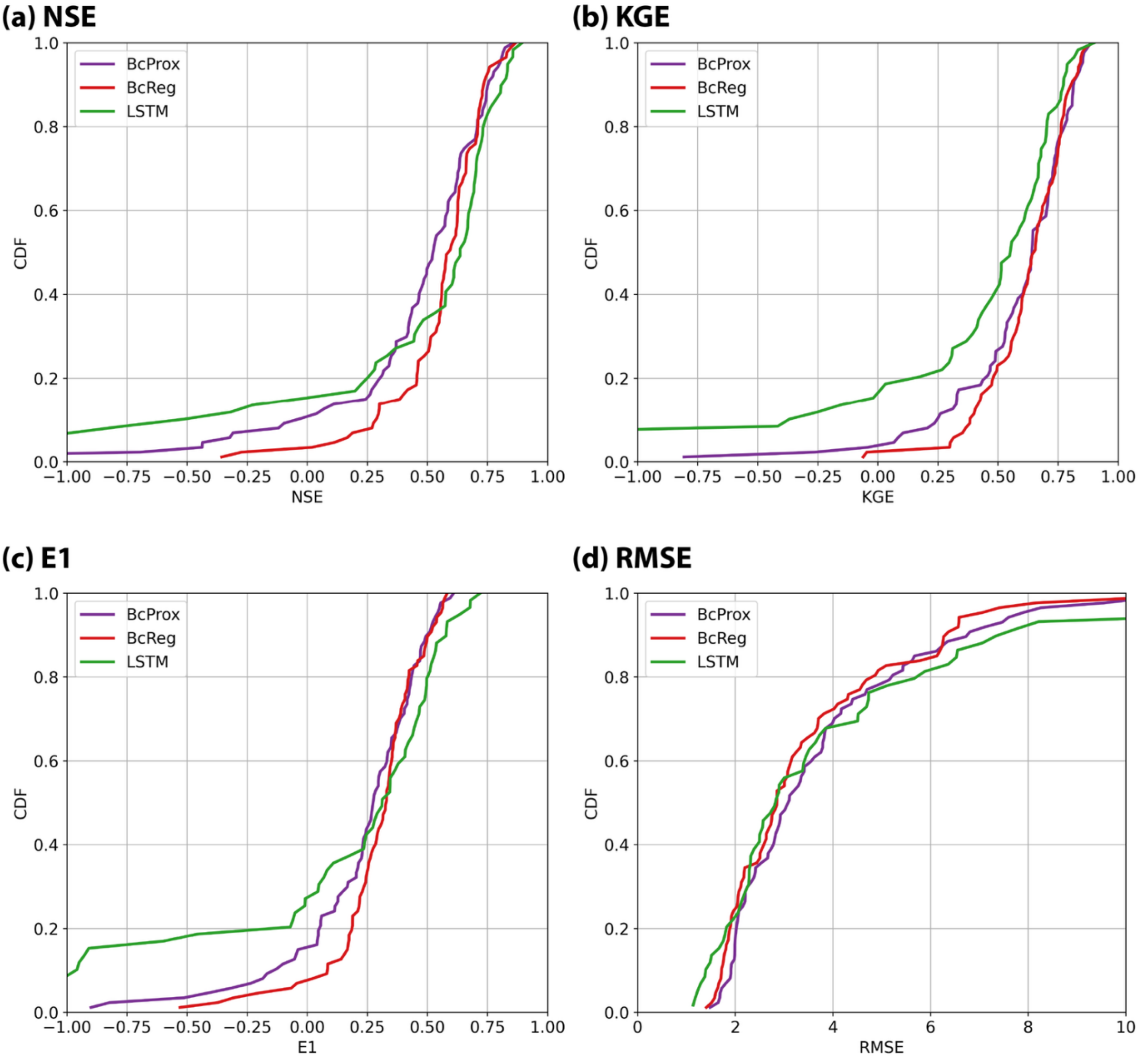

Figure 6 Model accuracy using (a) NSE, (b) KGE, (c) E1, and (d) RMSE using 12-fold cross validation. The results for HYPER-BcProx (purple), HYPER-BcReg (red), and LSTM (green) are shown.

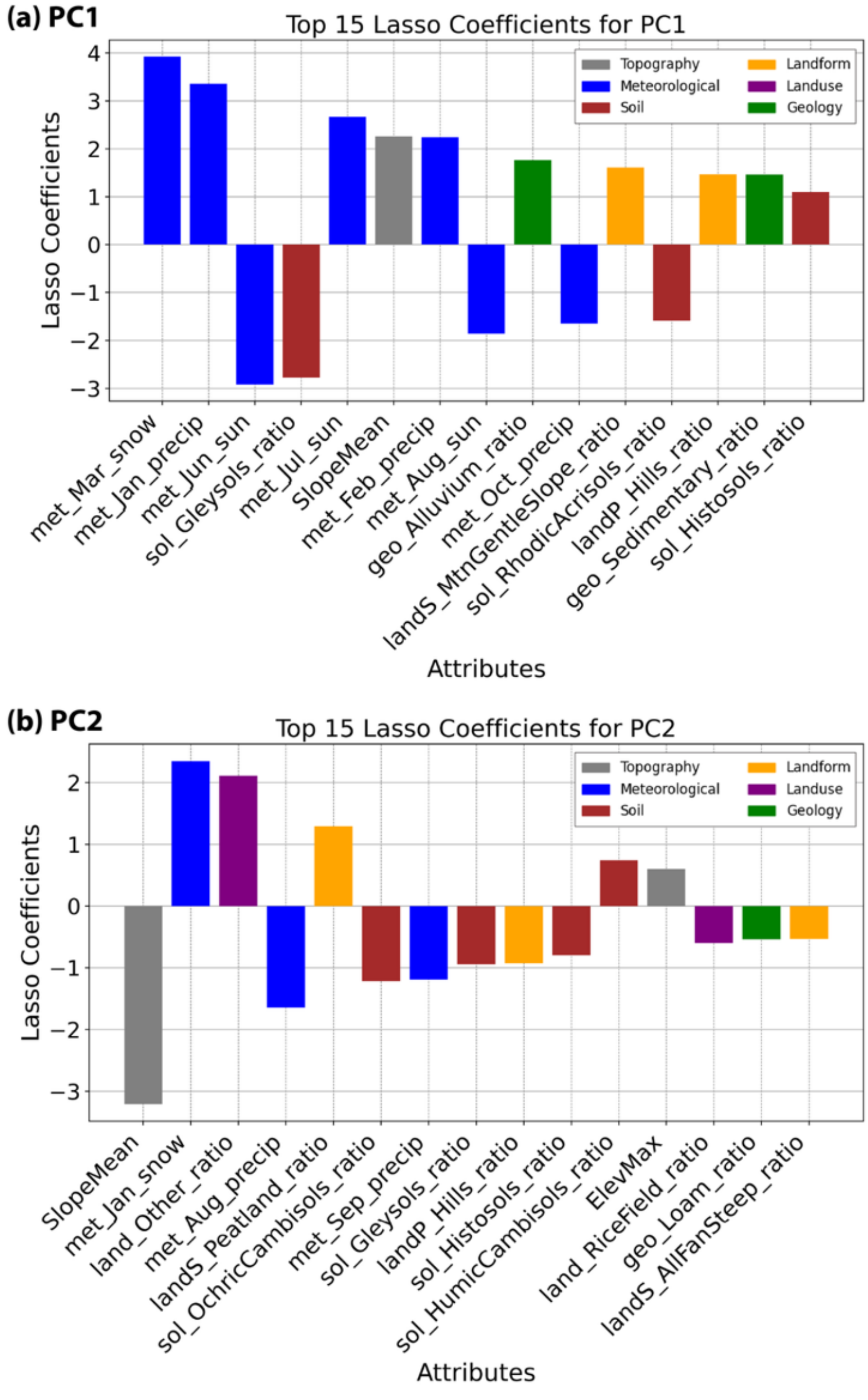

Figure 7 Lasso coefficient for each basin attribute for the first and second principal components in Japan. Due to visibility, only the 15 basin attributes with the largest absolute Lasso coefficient values are shown. The bars are colored based on the type of attributes, topography (gray), meteorological (blue), soil (brown), landform (yellow), land use (purple), and geology (green).

Table 2: Computational time required per basin when using HYPER-BcProx, HYPER-BcReg, and LSTM for calibration and prediction. * signifies the one time only prediction and BMA ensemble for the 47 MARRMoT models using MATLAB.

| | **Calibration** | **Prediction** |
|---|---|---|
| **HYPER-BcProx** | None | 3 min* + 5s |
| **HYPER-BcReg** | None | 3 min* + 6s |
| **LSTM** | 95 min | 6 s |

### 5.2.2 Experiment 3: Predictions in Data-Scarce Scenarios

Figure 8 shows the average prediction accuracy across 100 iterations for the 17 fixed "ungauged" basins using data from $n$ "gauged" basins. For each $n$, the mean prediction accuracy was calculated for each fixed basin, and the cumulative distribution functions (CDFs) were constructed using these mean values to summarize model performance across different gauging scenarios.

The HYPER methods show a clear positive relationship between the number of training basins and prediction accuracy. HYPER-BcReg consistently outperforms HYPER-BcProx, particularly in basins with lower predictability (the lower tail of the CDF). Notably, with 30 or more "gauged" basins, HYPER-BcReg achieves prediction accuracy in terms of NSE, RMSE, and E1 nearly equivalent to that observed in fully gauged basins (indicated by the black line in Fig. 8). For example, the median NSE value is 0.58 with 70 gauged basins and remains robust at 0.55 even with only 30 basins, closely approaching the 0.62 median NSE value achieved in the fully gauged experiment for the test basins. Table 3 summarizes the median values of NSE, KGE, E1, and RMSE across all models and gauging scenarios. This suggests that with a training set of approximately 30 basins out of 87 basins, HYPER-BcReg effectively captures the linkage between catchment characteristics and model weights, enabling reliable flood peak and hydrograph shape predictions in ungauged basins.

However, for the KGE metric, which emphasizes the full hydrograph shape and low-flow conditions, the prediction accuracy remains lower than for fully gauged basins. While HYPER-BcProx and HYPER-BcReg perform similarly in KGE (0.67 and 0.64, respectively, at n=70), they do not reach the gauged benchmark of 0.75. This indicates that while the relationship between basin attributes and model weights is well captured for flood peaks (NSE and RMSE), errors persist in simulating the comprehensive water balance components represented by KGE, reflecting the greater difficulty of generalizing bias and variability dynamics with limited data.

Furthermore, HYPER-BcReg and HYPER-BcProx show a faster saturation in predictive performance, reaching a plateau at approximately 20 gauged basins (Fig. 9). This demonstrates the two HYPER methods' efficiency in learning meaningful attribute-weight relationships with relatively fewer data.

When comparing to the LSTM model, even with 70 gauged basins, LSTM underperforms relative to both HYPER models in all metrics. Moreover, LSTM's prediction accuracy shows a counterintuitive decline as the number of gauged basins decreases from 70 to 30 or 20, yielding negative NSE values (-1.01 at n=30). This likely reflects LSTM's sensitivity to data quantity and variability, where, with moderate but insufficient data, LSTM may partially learn patterns but also overfit noise, increasing uncertainty and reducing generalization. The performance slightly recovers at extremely low data availability (n=3) because the model defaults to a simpler mean-like behavior rather than attempting to learn complex, unsupported dynamics.

Figure 9 shows the median and 10th to 90th percentile ranges of key evaluation metrics across 17 fixed "ungauged" basins from 100 random trials using different numbers of "gauged" basins. When examining the model prediction uncertainty, as the number of "gauged" basins decreases, HYPER-BcReg maintains stable accuracy and low uncertainty across most training basin sizes, with uncertainty only slightly increasing when fewer than 10 basins are gauged out of 87. HYPER-BcProx also shows relatively stable accuracy, though it exhibits slightly higher uncertainty and lower median accuracy than HYPER-BcReg, particularly in RMSE. In contrast, the LSTM's performance drops substantially, accompanied by a rapid increase in uncertainty (represented by the wide green shaded area).

Overall, these results demonstrate the robustness and data-efficiency of HYPER-BcReg in data-scarce environments, highlighting its advantage over LSTM and HYPER-BcProx in reliably transferring hydrological behaviors to ungauged regions using limited training networks.

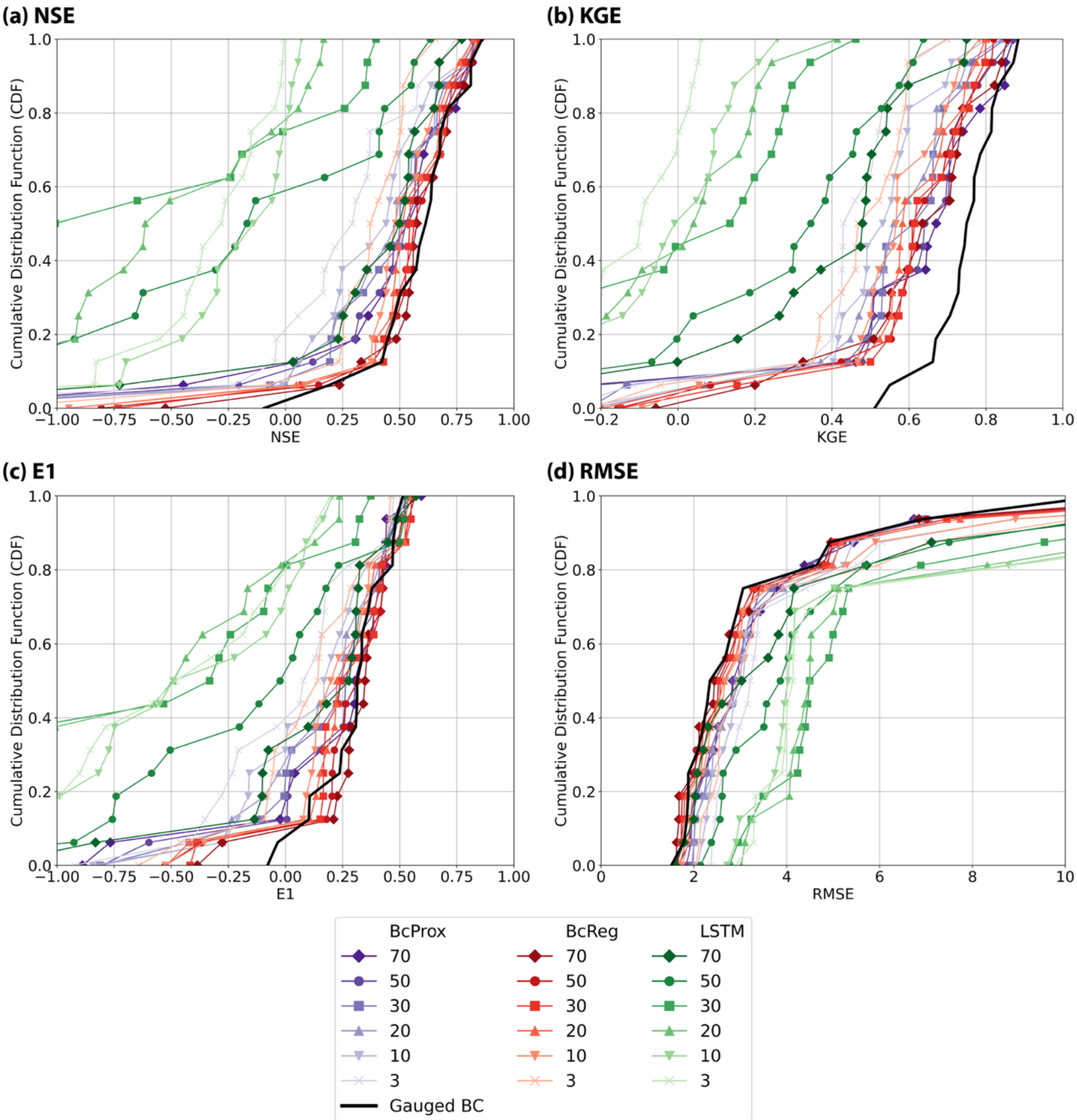

Figure 8 Cumulative distribution functions (CDFs) of model prediction accuracy for the 17 fixed "ungauged" basins, evaluated using (a) NSE, (b) KGE, (c) E1, and (d) RMSE. For each number of "gauged" basins, the mean prediction accuracy was first computed across 100 iterations for each fixed "ungauged" basin, and the CDF was then constructed using these 17 mean values. Models were trained on different number of "gauged" basins, ranging from 3, 10, 20, 30, 50, and 70 basins out of the total 87 basins. The results for HYPER-BcProx (purple), HYPER-BcReg (red), and LSTM (green) are shown with the darkest color showing the results when 70 basins were considered "gauged" and the lightest showing when only 3 basins were considered "gauged". The black line represents the benchmark performance when the testing basins are treated as gauged (Experiment 1).

Table 3: Median values of NSE, KGE, E1, and RMSE for HYPER-BcProx, HYPER-BcReg, and LSTM across different numbers of gauged basins (out of 87). Bolded values indicate the best performance for each metric at each number of gauged basins (highest for NSE, KGE, E1, and lowest for RMSE).

| **No. of Gauged Basins** | **HYPER-BcProx** | | | | **HYPER-BcReg** | | | | **LSTM** | | | |
|---|---|---|---|---|---|---|---|---|---|---|---|---|
| | **NSE** | **KGE** | **E1** | **RMSE** | **NSE** | **KGE** | **E1** | **RMSE** | **NSE** | **KGE** | **E1** | **RMSE** |
| 70 | 0.54 | **0.67** | 0.30 | 2.84 | **0.58** | 0.64 | **0.35** | **2.44** | 0.50 | 0.48 | 0.28 | 3.03 |
| 50 | 0.53 | **0.64** | 0.26 | 2.84 | **0.56** | 0.61 | **0.32** | **2.53** | -0.17 | 0.35 | -0.02 | 3.86 |
| 30 | 0.50 | 0.61 | 0.24 | 2.96 | **0.55** | **0.61** | **0.24** | **2.57** | -1.01 | 0.13 | -0.33 | 4.49 |
| 20 | 0.48 | 0.56 | **0.25** | 2.99 | **0.51** | **0.58** | 0.22 | **2.65** | -0.61 | 0.04 | -0.49 | 4.49 |
| 10 | 0.44 | 0.53 | 0.17 | 3.06 | **0.48** | **0.57** | **0.20** | **2.67** | -0.14 | -0.01 | -0.49 | 4.03 |
| 3 | 0.29 | 0.43 | 0.08 | 3.21 | **0.37** | **0.47** | **0.14** | **2.93** | -0.28 | -0.10 | -0.48 | 4.14 |

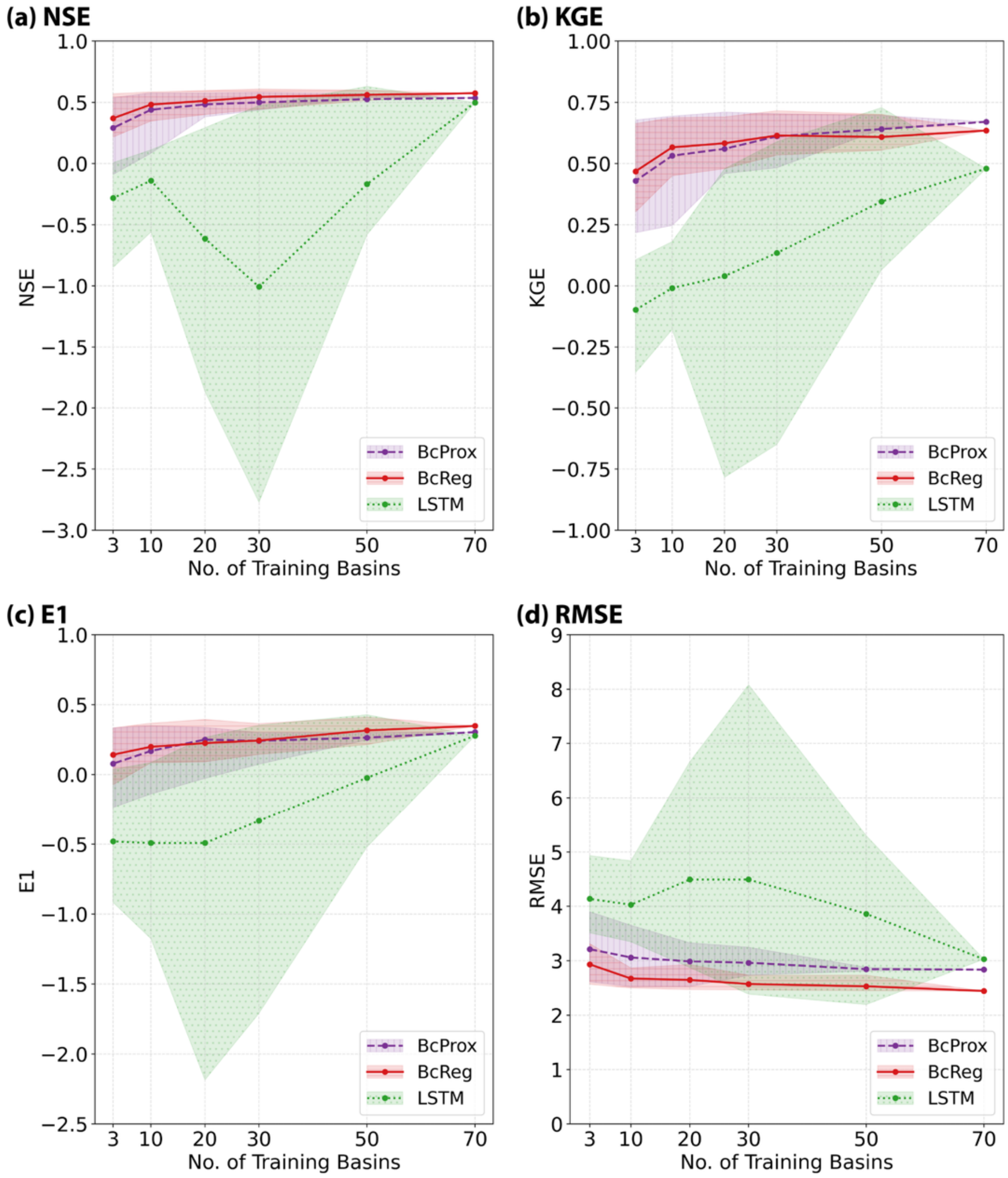

Figure 9 Model prediction uncertainty for the 17 fixed "ungauged" basins based on 100 random trials using different numbers of "gauged" basins, ranging from 3, 10, 20, 30, 50, and 70 basins out of the total 87 basins. They were evaluated based on (a) NSE, (b) KGE, (c) E1, and (d) RMSE. For each number of gauged basins, the lines represent the median of the evaluation metric across the 17 ungauged basins, while the shaded areas indicate the 10th to 90th percentile range of these median values over the 100 trials. The results for HYPER-BcProx (purple), HYPER-BcReg (red), and LSTM (green) are shown.

### 5.2.3 Experiment 4: Predictions in Remote Scenarios

Next, we evaluated the performance of HYPER-BcProx and HYPER-BcReg for predicting discharge in ungauged basins under remote scenarios, where all training basins were confined to the same geographic region. The results are shown in Fig. 10. The results were compared to when approximately the same number of basins (in this case, 15 basins) were selected randomly from all available basins (labeled as "Random").

Overall, for most regions, prediction accuracy was slightly lower than that of the randomly selected "gauged" basins, though the reduction was minimal. Comparing the two HYPER methods reveals distinct regional characteristics. While HYPER-BcReg demonstrated greater stability across all scenarios, HYPER-BcProx achieved equal or higher median accuracy in geographically central and western regions. Notably, for KGE and RMSE, HYPER-BcProx outperformed HYPER-BcReg when trained on North-Central, South-Central, or West basins. This suggests that in geographically homogeneous regions, comparable accuracy to non-remote scenarios can be achieved using simple spatial proximity, provided that the training basins adequately represent key catchment characteristics.

However, HYPER-BcProx showed lower accuracy in northern areas compared to HYPER-BcReg and especially significant accuracy degradation in terms of RMSE for Tohoku. A plausible explanation for these regional differences is geographic and climatic heterogeneity. Regions such as Hokkaido and Tohoku are located in northern Japan and experience heavy snowfall, which strongly influences hydrological processes. The concentration of snow-dominated basins within these regions complicates the generalization of spatial proximity model weights when applied to basins outside the snowy climate zones. In contrast, training basins from the West region, which has less snow influence, did not show a similar decrease in prediction accuracy for HYPER-BcProx, suggesting that climatic representativeness of training basins is crucial for reliable extrapolation. Importantly, HYPER-BcReg remained robust even in the northern regions, indicating that the regression-based approach effectively extracts generalizable physical relationships despite the regional bias.

When compared to the benchmark LSTM model, HYPER models demonstrated superior performance. The LSTM not only exhibited lower overall prediction accuracy but also a wider distribution of prediction errors across all four evaluation metrics, especially compared to HYPER-BcReg. The LSTM showed a high spread of NSE values in Tohoku, a high spread of RMSE in Hokkaido, and a large spread in KGE for South-Central, likely indicating overfitting to region-specific hydrological characteristics, such as snowmelt processes or distinct seasonal precipitation regimes, which reduce generalizability.

In summary, these results demonstrate that ungauged basin predictions using geographically constrained training data can reach accuracy comparable to randomly selected, non-remote gauged basins. While HYPER-BcProx is effective when the training region is climatologically representative (e.g., West/Central), HYPER-BcReg provides essential robustness when training data is confined to distinct peripheral regions (e.g., Hokkaido). This underscores the potential of regionally trained HYPER-BcReg models for discharge prediction beyond regional boundaries, based on the appropriate selection of representative training basins.

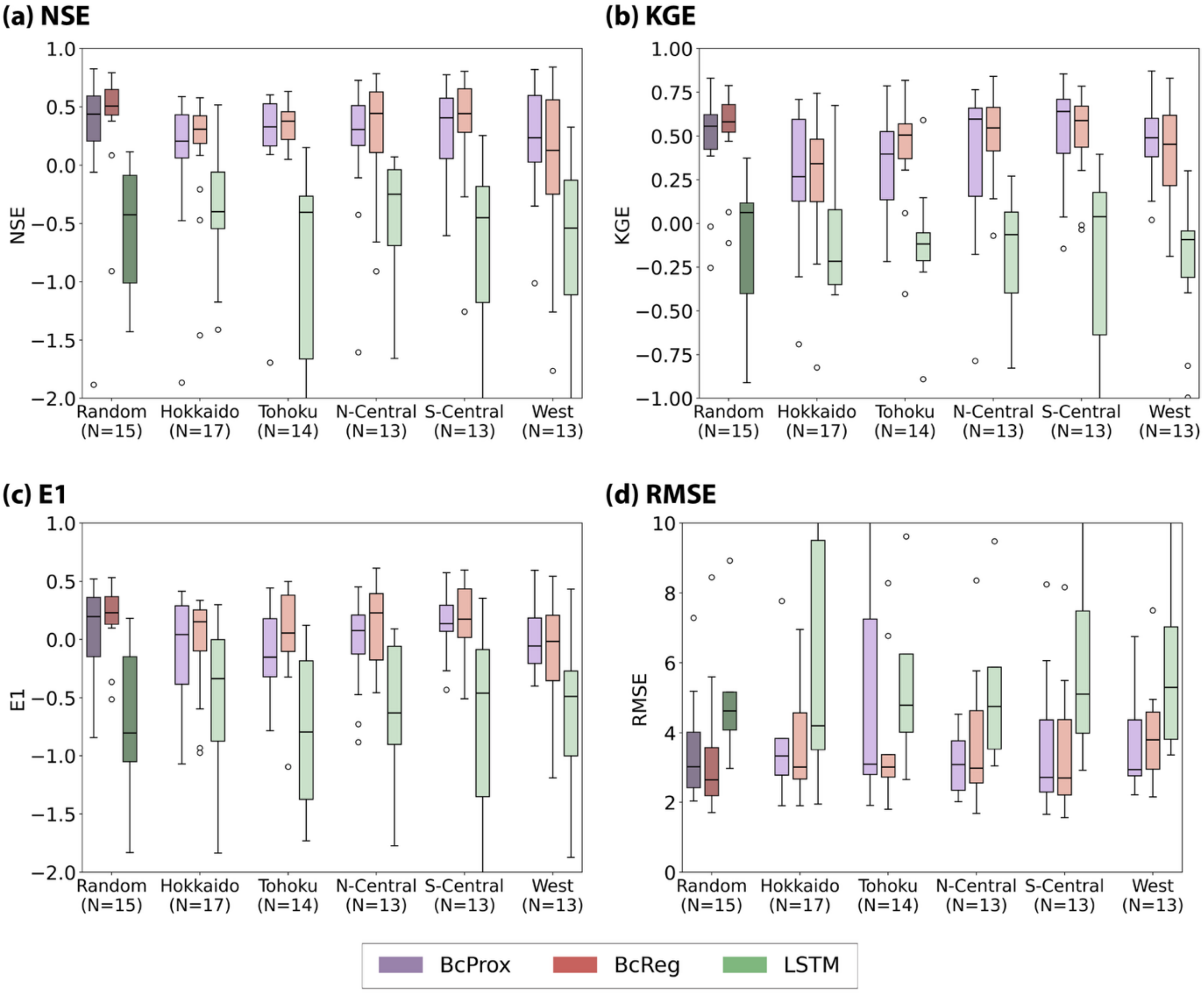

Figure 10 Model prediction accuracy of the 17 fixed “ungauged” basins using (a) NSE, (b) KGE, (c) E1, and (d) RMSE with “gauged” basins confined to a certain region (Hokkaido, Tohoku, North Central, South Central and West). The results for HYPER-BcProx (light purple), HYPER-BcReg (light red), and LSTM (light green) are shown for each region along with the results of Experiment 3 when the “gauged” basins is not confined to a certain region, labeled as “Random” shown for dark purple, dark red, and dark green respectively.

## 6 Conclusions & Discussion

This study presents a novel hybrid hydrological model, HYPER, that integrates multi-model ensembles with reservoir computing (RC). The HYPER model achieves a balance of high accuracy, interpretability, and low computational cost, making it particularly well-suited for global applications, including discharge predictions in ungauged basins. This section discusses the comparative performance of HYPER against prior studies, highlights its strengths and limitations, and suggests future research directions.

In terms of accuracy, the HYPER model consistently demonstrated strong performance across both gauged and ungauged basins. When applied to ungauged basin prediction, it matched or surpassed the accuracy of the LSTM benchmark model (Kratzert et al., 2019a). Even when the ratio of gauged basins against all basins was small, by leveraging ensemble diversity and the RC's ability to generalize from limited information, HYPER maintained low uncertainty while capturing essential flow dynamics.

Compared to traditional hydrological modeling efforts for prediction in ungauged basins, HYPER achieved competitive accuracy without the need for site-specific calibration. For instance, while the calibrated Variable infiltration capacity (VIC) land surface hydrological model yielded a median NSE of 0.44 across 531 U.S. basins (Liang et al., 1994; Tsai et al., 2021) and the Water And Snow balance MODeling system (WASMOD) model reported 0.61-0.82 in Norway (Yang et al., 2018), HYPER achieved intermediate yet competitive accuracy of 0.59 for data-rich scenario and 0.51 for data scarce scenario where only 20 out of 87 basins were "gauged". Importantly, this performance was maintained using entirely uncalibrated conceptual models, whereas many traditional models rely on calibration. Even in gauged basins, though not the primary target of the model, HYPER produced results comparable to the LSTM benchmark, especially in terms of the Kling-Gupta Efficiency (KGE), while showing only slight trade-offs in other metrics such as NSE, E1, and RMSE. Compared to AV-SWAT-X 2005 (NSE 0.56 in nine Xiangxi basins; Xu et al., 2010) and SAC-SMA (NSE 0.8 in the Leaf River Basin; Gupta et al., 1999), HYPER (NSE 0.62) outperformed the former and approached the latter's performance, even though SAC-SMA benefited from calibration within a single basin.

Alongside its accuracy, one of the most defining strengths of the HYPER model is its computational efficiency. Traditional hydrological models, particularly those applied to ungauged basins, often require $O(10^3 \sim 10^5)$ iterative simulations per basin, a process that is computationally prohibitive for global-scale applications. These processes are time-intensive and may still result in suboptimal predictions due to equifinality and data limitations. HYPER bypasses this entirely. By relying on a pre-defined ensemble and RC's closed-form linear regression training, it eliminates iterative parameter search. The reservoir computing component operates without iterative learning, and its compact architecture is well-suited for the relatively low-dimensional nature of rainfall-runoff processes. While initial hydrological model runs (e.g., MARRMoT) still require several minutes per basin, overall runtime is greatly reduced compared to conventional calibration-intensive approaches. Moreover, by filtering out unrealistic models, those that fail to reproduce basic discharge seasonality, and parallelizing the ensemble simulations, the overall computational burden can be further minimized, making the approach scalable for large regional or global applications.

Furthermore, the interpretability offered by the HYPER model is a key strength. Unlike "black-box" deep learning models such as LSTM, where input-output mapping is opaque, HYPER allows for physical insights into the modeling process. By applying PCA and Lasso regression to the weights, we identified that meteorological and topographical factors are the dominant predictors. This transparency not only supports scientific understanding but also enhances trust and usability in policy or operational contexts, where explainability is valued.

Despite the promising performance of the HYPER model, some limitations remain, particularly regarding its generalizability to hydrologically distinct regions. While both HYPER-BcReg and HYPER-BcProx generally performed well across Japan, HYPER-BcProx exhibited lower predictive accuracy when trained using only certain basins, especially those in snow-dominated northern regions. These issues mirror the findings of previous studies (Beck et al., 2016; Parajka et al., 2013), which point to the difficulties of modeling streamflow in snowmelt-dominated areas. In such regions, processes such as delayed snowmelt introduce non-linearities that are not fully captured by the current set of input features or the RC readout mechanism.

Part of this limitation may stem from the omission of key physical variables in the model. In its current implementation, the set of basin characteristics used for parameter regression is limited primarily to meteorological variables, topographic indices, and land use information. However, variables such as soil moisture, groundwater levels, and geological permeability can exert a strong influence on the rainfall-runoff relationship. Their absence may explain why HYPER-BcReg underperforms in regions with dominant subsurface flow or delayed runoff responses. Furthermore, while the RC component excels at capturing dynamic temporal patterns, its predictive performance remains constrained by the quality and representativeness of the upstream bias-corrected ensemble inputs. Another important aspect relates to the performance of the model's bias correction step. The current approach relies on a calibration-free hydrological model to estimate correction terms; however, in regions where the hydrological model is poorly suited, such as dry climates, this can introduce systematic errors. These errors propagate into the final predictions and may compromise HYPER's performance in such regions. Additionally, the evaluation of models in arid zones is frequently complicated by lower observational accuracy in gauging stations, where shifting rating curves or ephemeral flow conditions can introduce significant noise into the training data. Improving this bias correction component, for example, by incorporating hydrological models that do not require calibration yet better represent local hydrology, could significantly reduce such errors.

When comparing HYPER-BcProx and HYPER-BcReg, the regression-based method consistently outperformed the proximity-based method in terms of robustness. The superior performance of HYPER-BcReg can be attributed to its ability to assign unique weights to each basin based on its physical characteristics, rather than relying on the assumption that neighboring basins behave similarly. This finding stands in contrast to earlier studies (Merz and Blöschl, 2004; Petheram et al., 2012), where spatial proximity was shown to be a strong predictor of hydrological similarity. However, we argue that the higher performance of BcReg in our case is likely due to the large and diverse training dataset used in the regression process, which allowed it to learn more generalizable patterns across climatic and topographic gradients. Our findings also align with those from conceptual model studies (Ali et al., 2012), reinforcing the idea that regression-based transfer of model parameters can be effective, given sufficient and diverse training data.

A key consideration for the broader application of HYPER is its dependence on basin attributes. While this study utilized high-resolution national datasets from Japan, the framework is fully compatible with globally available standardized datasets, such as HydroATLAS (Linke et al., 2019). Preliminary assessments using global attributes indicate that the regression-based weight estimation remains robust, suggesting that the method is not limited to regions with bespoke national data. However, in this paper, the validation of HYPER was limited to basins within Japan, which, while diverse, may not fully reflect the hydrological variability found globally. To improve the model's transferability, it will be essential to train and test it on an expanded set of catchments across different climate zones, geological conditions, and hydrological regimes. In particular, incorporating basins with seasonal snow, glacier melt, or monsoonal rainfall patterns would provide a stronger test of the model's flexibility.

Moreover, testing the model in arid and intermittent rivers, which are underrepresented in the current dataset, could help to identify specific failure modes and inform model enhancements.

In summary, while some challenges remain, the HYPER model represents a significant step forward in rainfall-runoff prediction. By combining the computational efficiency and simplicity of reservoir computing with ensemble modeling and regression-based generalization, it performs competitively, if not better, than traditional hydrological or pure machine learning approaches. Its ability to deliver high accuracy with interpretable components and minimal calibration makes it particularly promising for both research and operational applications, especially in data-scarce regions.

## Code and Data Availability

The code for the HYPER model developed in this study, including scripts to reproduce all figures, is available at: https://doi.org/10.5281/zenodo.15676823. Hydrological modeling was performed using MARRMoT v1.3 (Knoben, 2019). Forcing and discharge data were obtained from MERV-Jp v2.0 (Sawada et al., 2022; Sawada and Okugawa, 2023). Basin characteristics (topography, soil, geology, landform, land use) were extracted from the National Land Information portal (Ministry of Land, Infrastructure, Transport and Tourism, 2025).

## Author Contribution

MF and YS designed the research. MF developed the methodology, conducted analyses, and drafted and edited the manuscript. YS supervised the project, reviewed the manuscript, and acquired funding.

## Competing Interests

The authors declare that they have no conflict of interest.

## Acknowledgements

This study was supported by the Foundation of River & basin Integrated CommunicationS (FRICS) grant, the JST Moonshot R&D Program (JPMJMS2281), and the KAKENHI grant (25H00760, 23K20967).

**References From the Supporting Information**